\documentclass[runningheads]{llncs}

\usepackage[mobile]{eccv}

\usepackage{eccvabbrv}

\usepackage{graphicx}
\usepackage{booktabs}
\usepackage{bm}
\usepackage{algorithm,algpseudocode}
\algrenewcommand\algorithmicrequire{\textbf{Input:}}
\algrenewcommand\algorithmicensure{\textbf{Output:}}
\DeclareMathOperator*{\argmin}{arg\min}

\usepackage[accsupp]{axessibility}  %
\newcommand{\circlenum}[1]{{\textcircled{\tiny{#1}}}}

\definecolor{cvprblue}{rgb}{0.21,0.49,0.74}
\usepackage[pagebackref,breaklinks,colorlinks,citecolor=cvprblue]{hyperref}

\usepackage{orcidlink}
\usepackage{float}

\begin{document}
\def\eg{\emph{e.g.}\@\xspace} \def\Eg{\emph{E.g.}\@\xspace}
\def\ie{\emph{i.e.}\@\xspace} \def\Ie{\emph{I.e.}\@\xspace}
\def\cf{\emph{c.f.}\@\xspace} \def\Cf{\emph{C.f.}\@\xspace}
\def\etc{\emph{etc.}\@\xspace} \def\vs{\emph{vs.}\@\xspace}
\def\etal{\emph{et al.}\@\xspace}
\def\wrt{\emph{w.r.t.}\@\xspace}
\title{Improving 2D Feature Representations by 3D-Aware Fine-Tuning} 

\titlerunning{Improving 2D Feature Representations by 3D-Aware Fine-Tuning}

\author{Yuanwen Yue\inst{1} \quad
Anurag Das\inst{2} \quad
Francis Engelmann\inst{1,3} \\ Siyu Tang\inst{1} \quad Jan Eric Lenssen\inst{2}}

\authorrunning{Y.~Yue et al.}

\institute{\textsuperscript{1} ETH Zurich \quad \quad \quad \quad  \textsuperscript{3} Google \\ \textsuperscript{2} Max Planck Institute for Informatics, Saarland Informatics Campus}

\maketitle
\begin{figure}
\vspace{-20px}
\centering
\includegraphics[width=0.99\textwidth]{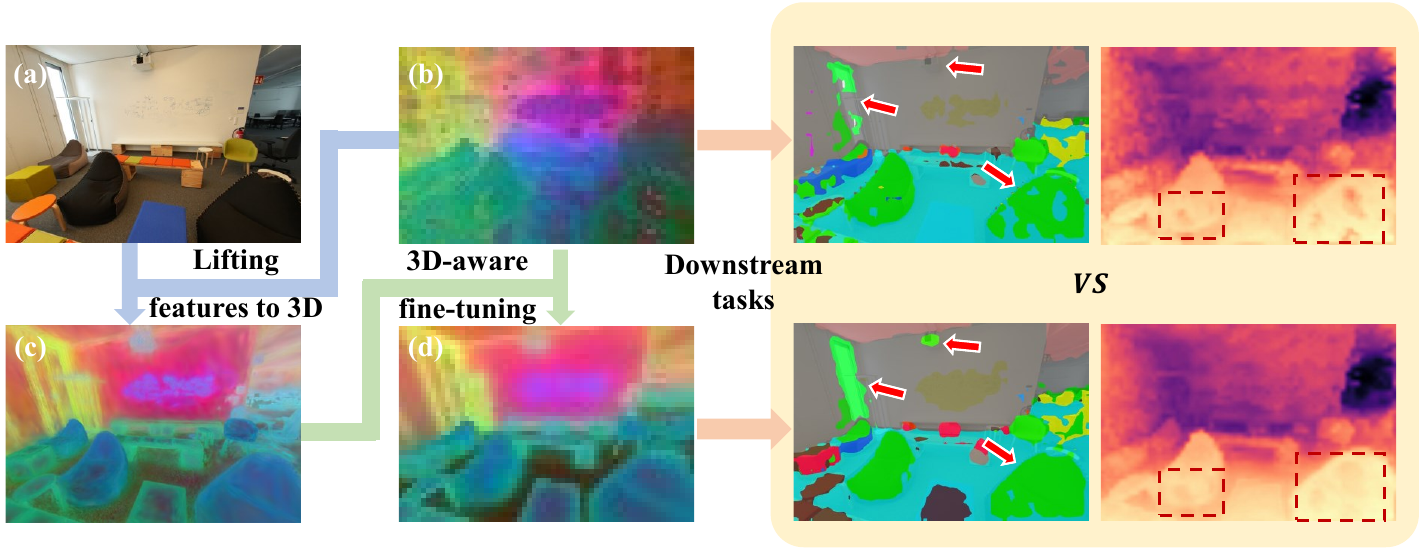}
    \caption{\small We propose 3D-aware fine-tuning to improve 2D foundation features. Our method starts with lifting \textit{2D image features} (\eg DINOv2~\cite{oquab2023dinov2}) \textbf{(b)} to a 3D representation. Then we finetune the 2D foundation model using the \textit{3D-aware features} \textbf{(c)}. We demonstrate that incorporating the \textit{fine-tuned features} \textbf{(d)} results in improved performance on downstream tasks such as semantic segmentation and depth estimation on a variety of datasets with simple linear probing \textbf{(right)}. Feature maps are visualized using principal component analysis (PCA).}
    \vspace{-7mm}
    \label{fig:teaser}

\end{figure}

\begin{abstract}
 Current visual foundation models are trained purely on unstructured 2D data, limiting their understanding of 3D structure of objects and scenes. In this work, we show that fine-tuning on 3D-aware data improves the quality of emerging semantic features. We design a method to lift semantic 2D features into an efficient 3D Gaussian representation, which allows us to re-render them for arbitrary views. Using the rendered 3D-aware features, we design a fine-tuning strategy to transfer such 3D awareness into a 2D foundation model. We demonstrate that models fine-tuned in that way produce features that readily improve downstream task performance in semantic segmentation and depth estimation through simple linear probing. Notably, though fined-tuned on a single indoor dataset, the improvement is transferable to a variety of indoor datasets and out-of-domain datasets. We hope our study encourages the community to consider injecting 3D awareness when training 2D foundation models. 
 Project page: \href{https://ywyue.github.io/FiT3D}{https://ywyue.github.io/FiT3D}.
\end{abstract}

\section{Introduction}
\label{sec:intro}
Ever since the emergence of deep neural networks, vision systems are largely trained on 2D datasets.
With the scalability of recent architectures, like vision transformers (ViT)~\cite{dosovitskiy2020image}, several large vision models~\cite{caron2021emerging,oquab2023dinov2,he2022masked,kirillov2023segment,rombach2022high} have been trained from a rich set of 2D images by either supervised or self-supervised learning.
Visual foundation models have shown impressive utility as general feature extractors that can be applied to improve results on downstream tasks, such as segmentation~\cite{li2023mask,tan2022semantic}, depth estimation~\cite{yang2024depth,saxena2024surprising,ke2023repurposing}, or correspondence estimation~\cite{amir2021deep,zhang2024tale}. 
They are trained on a large amount of readily available 2D images and, thus, learn statistics about object and scene structure in 2D-pixel space. 

Images, as a simple projection of our 3D world, are easy to obtain and provide an efficient way to depict the visual world while at the same time discarding explicit 3D geometry information. It is expected that vision systems purely trained on 2D images cannot fully understand the underlying 3D structure of our world~\cite{el2024probing}. There are several promising properties of our 3D world, for example, multi-view consistency, and multi-view fusion for solving single-view ambiguities. A crucial limitation of the training setups of these models is that they don't fully reason about the 3D structure of seen objects. Training images are presented to the network in an unstructured way, without any multi-view or video correspondences that would allow matching observations of the same object from multiple views. As a consequence, these models have limited 3D understanding of objects observed from, \eg, different views are not producing view-consistent features.

In contrast, when we humans observe images, we effortlessly achieve a holistic understanding by not only perceiving the 2D visual content but also exploiting the inferred underlying 3D structure, which we have learned through lifelong observation of stereo and temporal information. In this work, we investigate if large scale 2D vision models can also profit from equipping them with such 3D-aware understanding abilities induced by showing the right type of data.

\begin{figure}[t]
     \vspace{-10px}
\includegraphics[width=\columnwidth]{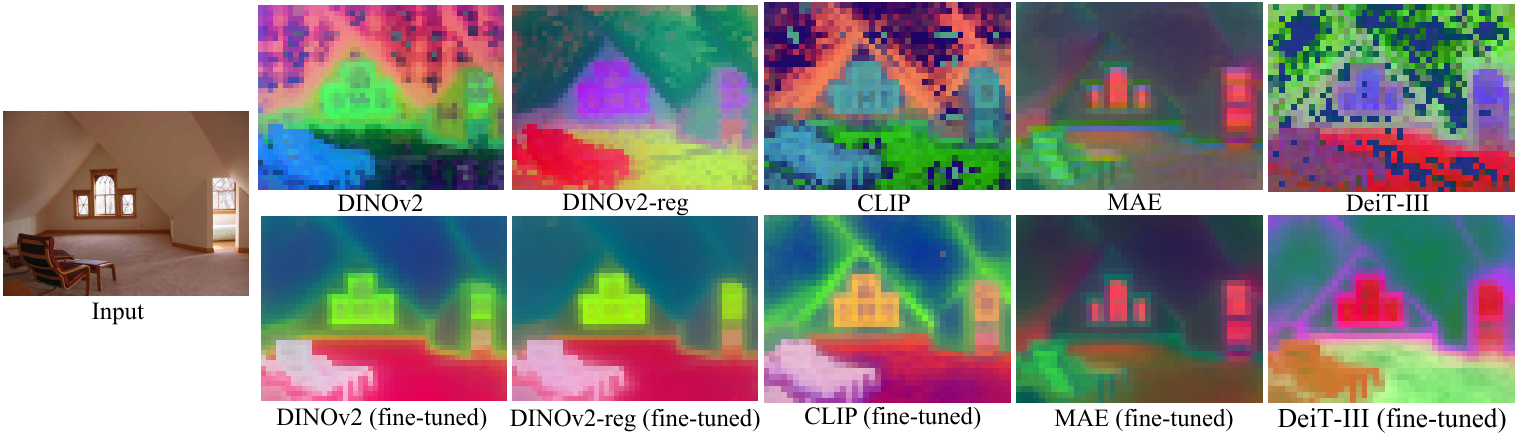}
    \vspace{-15px}
    \caption{Our 3D-aware fine-tuning is universal and applicable to a variety of 2D vision models, \eg DINOv2~\cite{oquab2023dinov2}, DINOv2-reg~\cite{darcet2023vision}, CLIP~\cite{radford2021learning}, MAE~\cite{he2022masked}, and DeiT-III~\cite{touvron2022deit} (\cf Sec.~\ref{sec:multi_models}).
    }
    \label{fig:addition_fig}
\end{figure}

To this end, we design a novel two-stage approach to improve the 3D-aware understanding ability of 2D foundation models. In the first stage, we aim to obtain 3D-aware features as training data. Motivated by recent advancements in neural scene representation, we design an approach to lift multi-view 2D foundation features into an efficient 3D Gaussian representation~\cite{kerbl20233d}. 
The lifting process exploits multi-view consistency and allows 2D features from different views to complement each other. Moreover, the fused features (Fig.~\ref{fig:teaser} (c)) exhibit high resolution with fine details thanks to the learned 3D structure, emerging from multi-view RGB guidance. Once trained, the 3D Gaussians can render features for arbitrary views. In the following, we refer to features obtained in this way as 3D-aware.
 In the second stage, we utilize the rendered 3D-aware features to finetune the 2D foundation models (Fig.~\ref{fig:addition_fig}). To this end, we design an efficient fine-tuning strategy to transfer such 3D awareness into 2D foundation models.
After fine-tuning, we evaluate the feature quality on downstream tasks that might profit from a better 3D understanding, namely semantic segmentation and depth estimation. Extensive experiments demonstrate that incorporating the 3D-aware features improves downstream tasks with simple linear probing and exhibits generalization ability on out-of-domain datasets.

\section{Related Work}
\label{sec:related_work}
We give an overview about recent self-supervised 2D representation learning techniques in Sec.~\ref{sec:2d_rep_learning}, and how emerging features have been distilled into 3D representations in Sec.~\ref{sec:distilling_features}. Then, we discuss previous work that utilizes 3D information to improve 2D representation methods in Sec.~\ref{sec:3d-to-2d}.
\subsection{2D Representation Learning} 
\label{sec:2d_rep_learning}
Representation learning~\cite{bengio2013representation} has achieved remarkable progress in the image domain. It aims to learn generalizable visual features from a rich set of data. Self-supervised representation learning has gained particular interest since it does not require labeled data. Early works employ pretext tasks for pre-training, which aim to exploit inherent data attributes to automatically generate surrogate labels~\cite{dosovitskiy2014discriminative,gidaris2018unsupervised,noroozi2016unsupervised,doersch2015unsupervised,wang2015unsupervised,pathak2017learning}. Later, contrastive learning~\cite{hadsell2006dimensionality} has been popularly used for representation learning by leveraging discriminative
signals between images or groups of images~\cite{caron2021emerging,oquab2023dinov2,he2020momentum,caron2020unsupervised,grill2020bootstrap}. More recently, motivated by BERT~\cite{devlin2018bert}, a new paradigm of masked image modeling~\cite{chen2020generative,bao2021beit,he2022masked} has been proposed for scalable visual learning. Nevertheless, all those methods are only trained on 2D image data, without accessing the underlying 3D structure. Our work aims to supplement the features purely learned from 2D observations with 3D awareness. 
 
\subsection{Distilled Feature Fusion Fields}
\label{sec:distilling_features}
Neural radiance fields (NeRF)~\cite{mildenhall2021nerf} emerge as a promising scene representation for high-quality 3D reconstruction and novel view synthesis. 
Recently, some works \cite{tschernezki2022neural,kobayashi2022decomposing,goel2023interactive,kerr2023lerf, engelmann2024opennerf} explore distilling pre-trained image features (\eg DINO~\cite{caron2021emerging}, CLIP~\cite{radford2021learning}, LSeg~\cite{li2022languagedriven}, or OpenSeg~\cite{ghiasi2022scaling}) into NeRF via neural rendering. Without requiring any labels, such distilled feature fusion fields enable several zero-shot 3D scene understanding tasks, \eg segmentation, scene editing, and open-vocabulary queries. We share similar inspiration from these works by distilling 2D features into a 3D representation. However, instead of focusing on perception tasks with feature fields, 
we are interested in leveraging the rendered 3D-aware features to in turn improve the 2D feature extractor. We demonstrate that the transferred 3D awareness can readily improve the 2D features on both semantic and geometric tasks. Moreover, 
we extend the recent Gaussian-based representation~\cite{kerbl20233d} by designing a method to distill 2D features into 3D Gaussians while keeping high efficiency and memory under bound. There are several concurrent works introducing 3D Gaussians with semantic features \cite{Qin_2024_CVPR, shi2024language, zhou2024feature}. However, none of these works distill features back into 2D models. Our work shows, for the first time, that semantic features fused into 3D representations can effectively improve 2D foundation models via fine-tuning.

\subsection{Injecting 3D Priors to 2D}
\label{sec:3d-to-2d}
Existing works mainly focus on fusing multi-view 2D features into the 3D representation~\cite{ha2022semantic,Weder2024labelmaker,Huang2023Segment3D,mazur2022feature,peng2022openscene,Takmaz2023NIPS,jatavallabhula2023conceptfusion,shen2023distilled}. Little attention has been paid to the other direction of incorporating 3D awareness into 2D representation learning. Pri3D~\cite{hou2021pri3d} uses geometric constraints (multi-view consistency and 2D-3D correspondence) from RGB-D reconstructions to learn 3D
priors for image-based representations with contrastive learning. Recently, inspired by the masked
autoencoder (MAE)~\cite{he2022masked}, several works adopt the masked image modeling strategy to learn 3D priors~\cite{bachmann2022multimae, hou2023mask3d, weinzaepfel2022croco}. 
However, all these methods require pre-training the 2D feature extractor, typically a Vision Transformer (ViT) backbone~\cite{dosovitskiy2020image}, using their hand-crafted pretext tasks. The pre-trained models are then employed to downstream tasks via fine-tuning. 
By contrast, we aim to transfer the 3D awareness embedded in multi-view fused features to the 2D feature extractor through fine-tuning with little computational resources. Our 3D-aware features readily improve downstream task performance with simple linear probing. In addition, we find our 3D-aware features exhibit cleaner and more detailed feature maps compared with the original 2D features (see Sec.~\ref{sec:features} in appendix), while several concurrent works specifically denoise or sharpen 2D feature maps~\cite{yang2024denoising,fu2024featup}.

\section{Method}
\label{sec:method}
\begin{figure}[t]
\centering
\includegraphics[width=1\textwidth]{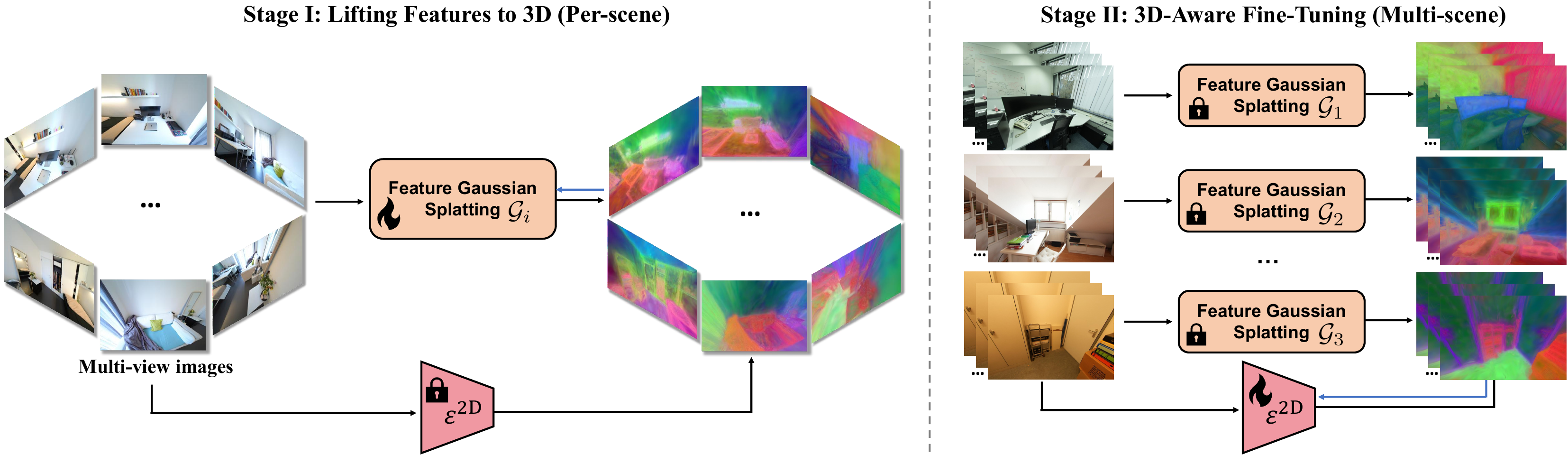}
    \caption{\textbf{Overall pipeline.} We present a two-stage pipeline. In the first stage, we lift 2D foundation features (\eg DINOv2~\cite{oquab2023dinov2}) into 3D-aware features by training 3D Gaussian representation $\mathcal{G}_{i}$. In the second stage, we use the rendered features to finetune the 2D foundation model $\varepsilon ^{2D}$. With \color{blue}{$\rightarrow$} \color{black} we denote gradient flow.}
    \label{fig:overview}
\end{figure}
In this section, we introduce our method for fine-tuning 2D foundation models with 3D-aware features. We present a two-stage pipeline (\cf Fig~\ref{fig:overview}). In the first stage, we lift per-view 2D features into a multi-view consistent and 3D-aware representation. The representation and setup are described in Sec.~\ref{sec:lifting_features}. In the second stage, we use the obtained 3D-aware feature representations as training dataset to finetune the 2D feature extractor, which is detailed in Sec.~\ref{sec:finetuning}. 
Last, we describe the linear probing methodology for feature evaluation in Sec.~\ref{sec:linear_probing}.
\begin{figure}[t]
\centering
\includegraphics[width=1\textwidth]{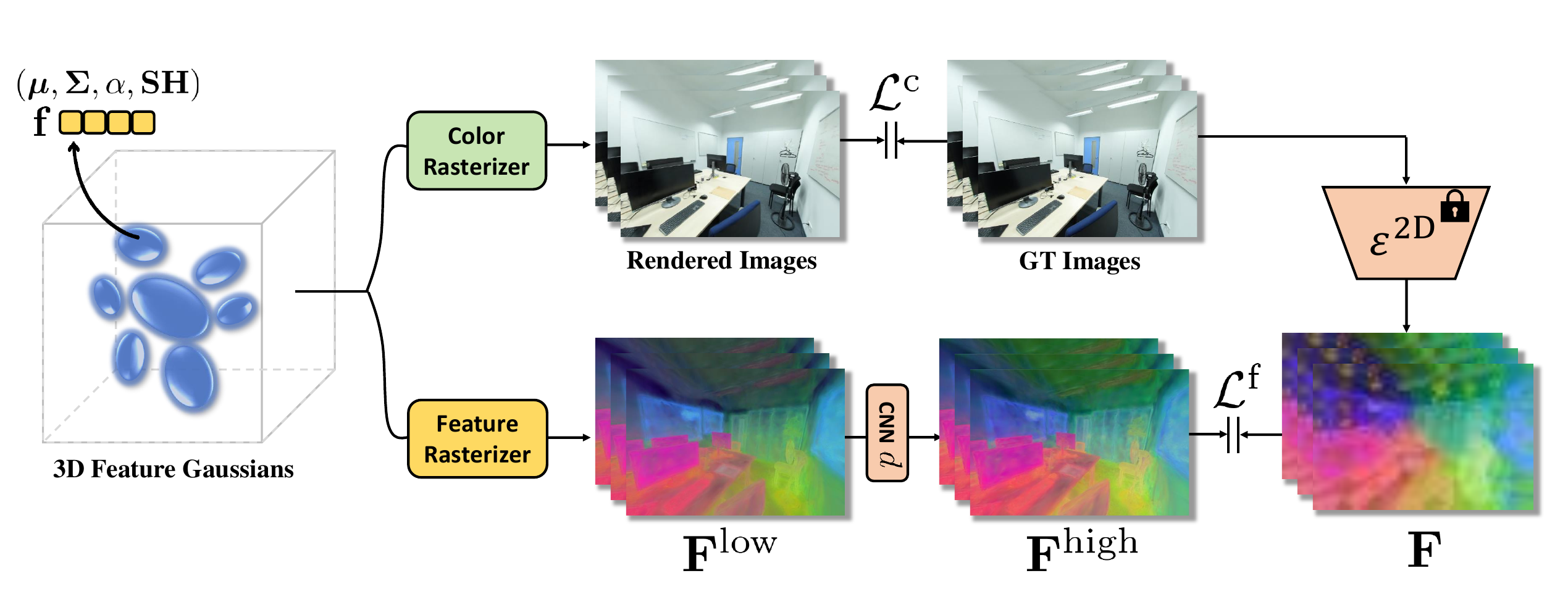}
    \caption{\small \textbf{Lifting 2D features into 3D Gaussian representation.} We equip each Gaussian with a low-dimensional feature vector $\mathbf{f}$. We render colors using the same color rasterizer as Gaussian splatting~\cite{kerbl20233d}. We design a feature rasterizer to render a low-dimensional feature image $\mathbf{F}^{\mathrm{low}}$, which is subsequently projected to a high-dimensional feature image $\mathbf{F}^{\mathrm{high}}$ using a simple CNN. We use 2D foundation features $\mathbf{F}$ from model $\varepsilon ^{\mathrm{2D}}$ to supervise the feature learning.}
    \label{fig:feature_gaussians}
\end{figure}

\subsection{Lifting Features to 3D}
\label{sec:lifting_features}
Lifting semantic 2D features into 3D has been a trend recently and several different options exist (\cf Sec.~\ref{sec:related_work}). For our purposes of using larger amounts of scenes as training data for 2D models, the most important aspect is efficiency. The representation needs to  (1) be able to efficiently fit a large number of scenes into 3D representations and (2) have a fast rendering mechanism for efficient integration into a fine-tuning loop of a 2D foundation model. Thus, we utilize the recent advances in 3D Gaussian splatting~\cite{kerbl20233d}, which enable fast optimization and real-time rendering. Fig.~\ref{fig:feature_gaussians} illustrates how we extend Gaussian splatting to lift 2D foundation features and we detail the method below.

\vspace{0.1cm}
\noindent\textbf{3D feature Gaussians.} Adapting the formulation of 3D Gaussian splatting~\cite{kerbl20233d}, we define a set of 3D Gaussians as
\begin{equation}
    \mathcal{G} = \{(\bm{\mu}, \mathbf{s}, \mathbf{R}, \alpha, \mathbf{SH}, \mathbf{f})_j)\}_{1\leq j \leq M} \textnormal{,}
\end{equation}
where $\bm{\mu} \in \mathbb{R}^3$ is the 3D mean of the Gaussian, $\mathbf{S} = \textnormal{diag}(\mathbf{s}) \in \mathbb{R}^{3\times 3}$ is the Gaussian scale, $\mathbf{R}\in \mathbb{R}^{3\times 3}$ its orientation, $\alpha \in \mathbb{R}$ is a per-Gaussian opacity, and $\mathbf{SH}$ a vector of spherical harmonic coefficients, encoding view-dependent color. The Gaussian covariance matrix is obtained by combining scale and orientation as $\mathbf{\Sigma} = \mathbf{R}\mathbf{S}\mathbf{S}^\top\mathbf{R}^\top$. In addition to the original parameters, we introduce a per-Gaussian feature vector $\mathbf{f}\in \mathbb{R}^D$ to store distilled 2D features in 3D space. Those feature vectors are rasterized into a 2D feature image with our designed feature rasterizer. Inspired by the differentiable color rasterizer of Gaussian splatting, we rasterize the features using point-based $\alpha$-blending as follows:
\begin{equation}
\mathbf{F}^{\mathrm{low}} = \sum_{i\in \mathcal{N}}\mathbf{f}_{i}\alpha _{i}\prod_{j=1}^{i-1}(1-\alpha _{i})
\end{equation}
where $\mathcal{N}$ is a set of ordered Gaussians overlapping the pixel, $\mathbf{f}_{i}$
is the feature of each Gaussian and $\alpha _{i}$ is given by evaluating a
2D Gaussian with covariance $\mathbf{\Sigma}$ multiplied with a
learned per-point opacity.

\vspace{0.1cm}
\noindent\textbf{Up-projecting features.} A strong limitation of 3D Gaussians as representation is their memory consumption. Since there can be millions of Gaussians per scene, it is impossible to store, \eg, the $384$-dimensional DINO features directly on each of the 3D Gaussians. Therefore, to stay memory efficient and keep the fast rendering process, we opt for storing lower dimensional features $\mathbf{f} \in \mathbb{R}^D$ with $D$ \(<\!\!<\) $384$ and train a scene-specific pixel-space CNN decoder $d: \mathbf{F}^\textnormal{low} \mapsto \mathbf{F}^\textnormal{high}$ to up-project feature images into high-dimensional feature space after rendering. We analyze the trade-off introduced by this approach in Sec.~\ref{subsec:ablation}.

\vspace{0.1cm}
\noindent\textbf{Optimization.} For a given scene, the full 3D Gaussian representation, including our distilled features, is obtained using optimization.
Let $\{\mathbf{I}_i\}_{1\leq i \leq N}$ be a set of multi-view images of a scene with corresponding camera parameters, $\{\mathbf{F}_i\}_{1\leq i \leq N}$ a corresponding set of feature maps from a 2D feature extractor (\eg DINOv2~\cite{oquab2023dinov2}), and $r^\textnormal{rgb}$, $r^\textnormal{feat}$ rasterization functions that render a set of Gaussians into an RGB or feature image, respectively, using the camera pose $\mathbf{P}_{i}$ of image $i$. Then, we optimize the Gaussian parameters, to optimally represent images $\mathbf{I}_i$ and feature images $\mathbf{F}_i$:
\begin{equation}
    \hat{\mathcal{G}} = \argmin_{\{(\bm{\mu}, \mathbf{s}, \mathbf{R}, \alpha, \mathbf{SH}, \mathbf{f})_i\}} \sum_{i=1}^N\mathcal{L}^c(r^\textnormal{rgb}(\mathcal{G}, \mathbf{P}_{i}),\mathbf{I}_i) + \mathcal{L}^f(d(r^\textnormal{feat}(\mathcal{G}, \mathbf{P}_{i}),\mathbf{F}_i) \textnormal{,}
\end{equation}
where $\mathcal{L}^c$ is a pixel-wise $l_{1}$ loss combined with a D-SSIM term on RGB images, and $\mathcal{L}^f$ is a pixel-wise $l_{1}$ loss on feature images. Notably, we only optimize $\mathbf{f}$ with gradients coming from $\mathcal{L}^f$ (feature images) and the rest of the parameters only on $\mathcal{L}^c$ (RGB loss). This has proven to be essential to obtain a consistent 3D feature representation, as a loss from feature space does not lead to correct Gaussian mean, covariance and opacity. We speculate that the reason for this is the missing 3D consistency of the 2D feature extractor. Only through forcing them into a 3D consistent representation, we make them consistent in return.

\begin{algorithm}[t]
\caption{3D-aware fine-tuning algorithm}
    \begin{algorithmic}[1]
    \Require Pre-trained Feature Gaussian representations $\{\mathcal{G}_1, ..., \mathcal{G}_K\}$, pre-trained 2D feature extractor
    $\varepsilon_\theta^{2D}$, a set of images $\left \{ \mathbf{I}_{i} \right \}_{i=1}^{N}$ and associated camera poses $\left \{ \mathbf{P}_{i} \right \}_{i=1}^{N}$.
    \Ensure Fine-tuned 2D feature extractor $\varepsilon_{\hat{\theta}}^{2D}$.
    \State Load $\mathcal{G} \sim \{\mathcal{G}_1, ..., \mathcal{G}_K\}$
    \While {fine-tuning}
    \State Sample an image $\mathbf{I}_{i}$ and camera pose $\mathbf{P}_{i}$, $\quad i\sim \mathcal{U}\{1,N\}$
    \State Retrieve associated feature Gaussian $\mathcal{G}$ and CNN decoder $d$
    \State Render $\mathbf{F}^{\mathrm{high}} \leftarrow d(r^\textnormal{feat}(\mathcal{G}, \mathbf{P}_{i})$)
    \State Step $\theta$ by minimizing $\mathcal{L}(\varepsilon_\theta^{2D}(\mathbf{I}_{i}), \mathbf{F}^{\mathrm{high}})$
    
    \EndWhile
    
    \hspace{-0.65cm}\Return $\varepsilon_{\hat{\theta}}^{2D}$
    \end{algorithmic}
    \label{alg:finetune} 
\end{algorithm}
\subsection{3D-Aware Fine-Tuning}
\label{sec:finetuning}
The procedure described in the last section is used to fit 3D feature Gaussian representations of $K$ scenes. 
The algorithm of 3D-aware fine-tuning is outlined in Algorithm~\ref{alg:finetune}. 
The fine-tuning process requires training pairs of original 2D feature maps and 3D-aware feature maps. Since it is memory-intensive to save the feature maps, we generate the training pairs on the fly. 
Considering it is time-consuming to load each pre-trained Gaussian when rendering features, we pre-load all the Gaussians into CPU memory. 
In each step of the training loop, we randomly sample a view from all the training images, then retrieve its associated feature Gaussian and scene-specific CNN decoder and finally render features $\mathbf{F}^\mathrm{high}$ as the ground truth features for fine-tuning. The fine-tuning loss is a $l_{1}$ loss between $\mathbf{F}^\mathrm{high}$ (resized) and the output features of the fine-tuned 2D feature extractor. 

The above design makes the fine-tuning process efficient and keeps memory consumption under control. Notably, we only need to fine-tune the 2D feature extractor with a small number of epochs (\eg 1 epoch for DINOv2~\cite{oquab2023dinov2}) with a small learning rate without additionally introducing any network component. The fine-tuning process is fast and computation-friendly. An analysis of fine-tuning time is in Sec.~\ref{subsec:ablation}.

\subsection{Linear Probing for Downstream Tasks}
\label{sec:linear_probing}
After fine-tuning on 3D-aware features, we evaluate the emerging features on a set of standard benchmark downstream tasks. To this end, we train a linear head on top of the features to solve tasks of semantic segmentation and depth estimation on several datasets. 

\vspace{0.1cm}
\noindent\textbf{Semantic segmentation.}  A linear layer is trained to predict class logits from patch tokens. The linear layer produces a low-resolution logit map, which is then upsampled to full resolution to obtain a segmentation map.

\vspace{0.1cm}
\noindent\textbf{Depth estimation.} We concatenate the \texttt{[CLS]} token of the ViT to each patch token. We divide the depth prediction range into 256 uniformly distributed bins~\cite{bhat2021adabins} and use a linear normalization. Then a simple linear layer is trained using a classification loss.

\vspace{0.1cm}
\noindent\textbf{Feature assembly.} We concatenate original 2D features with our fine-tuned features. We observe this is key to preserving the generalization ability of the original 2D feature extractor while incorporating the 3D awareness in our fine-tuned features. Different strategies for feature assembly are evaluated in Sec.~\ref{subsec:ablation}. 

\section{Experiments}
\label{sec:experiments}
\subsection{Datasets}
\label{sec:datasets}
\textbf{Training.} We train the feature Gaussians on ScanNet++~\cite{yeshwanth2023scannet++}, which is a large-scale dataset of 3D indoor scenes containing sub-millimeter resolution laser scans, registered DSLR images, and commodity RGB-D streams from iPhone. We train on the official training split of 230 scenes, which contain 140451 views.

\vspace{0.1cm}
\noindent\textbf{Evaluation.} To examine the effectiveness of the fine-tuned features, we conduct extensive experiments on downstream 2D tasks including semantic segmentation and depth estimation. There is no direct competitor in our study and we instead focus on whether our 3D-aware fine-tuning can bring performance gains compared with the standard 2D feature extractor. We conduct most of the experiments with DINOv2~\cite{oquab2023dinov2} while also demonstrating the universality of our approach with other vision models in Sec.~\ref{sec:multi_models}. We first evaluate on ScanNet++~\cite{yeshwanth2023scannet++} validation set, which contains 50 scenes with 30638 images. Then we move on to other indoor datasets ScanNet~\cite{dai2017scannet} and NYUd~\cite{silberman2012indoor}.
which have a similar data distribution with ScanNet++ but were captured with different sensors. 
To investigate the generalization ability of the fine-tuned features, we also perform out-of-domain evaluation on generally distributed datasets including ADE20k~\cite{zhou2017scene}, Pascal VOC~\cite{everingham2015pascal} and the outdoor dataset KITTI~\cite{geiger2013vision}.
\subsection{Implementation Details}
\label{sec:implementation_details}
\textbf{Feature Gaussians.} We wrote custom CUDA kernels for feature rasterization. Each Gaussian is initialized with a random feature vector with a dimension of 64. We implement the up-projecting CNN with a single convolutional layer with a kernel size of 3×3. We train the feature Gaussians of each scene for novel view synthesis and feature rendering jointly for 30000 iterations. 

\vspace{0.1cm}
\noindent\textbf{Fine-tuning.} We finetune DINOv2 small with a feature dimension of 384 with a batch size of 2 with a learning rate of 1e-5 for 1 epoch.  We use horizontal flip as data augmentation. We use the AdamW~\cite{loshchilov2017decoupled} optimizer with a weight decay factor 1e-4. The fine-tuning on a single Nvidia Tesla A100 takes 8.5 hours.

\vspace{0.1cm}
\noindent\textbf{Linear probing.} We follow the linear probing protocol with DINOv2~\cite{oquab2023dinov2} to ensure a fair comparison. For semantic segmentation, we train the linear layer for 40K iterations with 8 GPUs. For depth estimation, we train the linear layer for 38400 iterations with 8 GPUs. In addition, we use the same data augmentation and learning rate schedule with DINOv2.

\subsection{Within-domain Evaluation}
\label{sec:indomain}

\vspace{0.1cm}
\noindent\textbf{Quantitative comparison.}
We demonstrate the effectiveness of incorporating our 3D-aware features on downstream semantic segmentation (see Tab.~\ref{tab:indoor_sem}) and depth estimation task (see Tab.~\ref{tab:indoor_depth}) for indoor scenes. For semantic segmentation task, our 3D aware features consistently improve DINOv2 features, achieving a significant performance gain of 2.6\%, 2.0\% mIoU, and 1.2\% on ScanNet++~\cite{yeshwanth2023scannet++}, NYUv2~\cite{silberman2012indoor} and ScanNet~\cite{dai2017scannet} datasets, respectively. Our 3D-aware DINOv2 features also improve performance on the depth estimation task. In particular, our enhanced features consistently reduce the RMSE across datasets by achieving 0.34 \vs 0.37 (DINOv2) for ScanNet++~\cite{yeshwanth2023scannet++}, 0.42 \vs 0.44 (DINOv2) for NYUv2~\cite{silberman2012indoor} and 0.29 \vs 0.31 (DINOv2) for ScanNet~\cite{dai2017scannet} datasets.

\begin{figure}[t!]
\centering
\includegraphics[width=\textwidth]{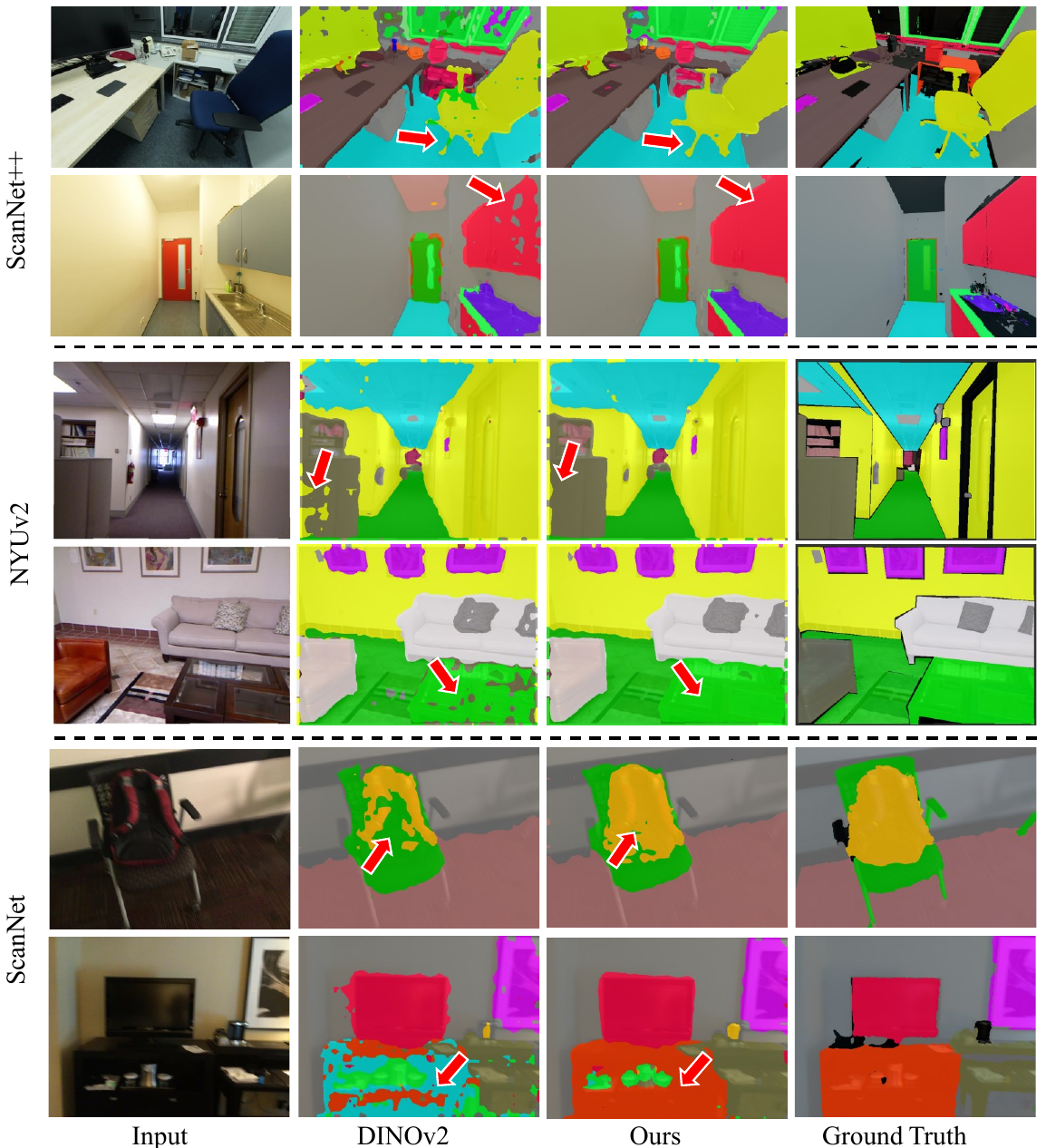}
    \caption{\small \textbf{Semantic segmentation on indoor datasets with linear probing}. Incorporating our 3D-aware fine-tuned features helps obtain cleaner and more compact segmentation results, especially for detailed structures and in homogeneous regions.}
    \label{fig:indoor_sem}
\end{figure}

\begin{figure}[h!]
\centering
\includegraphics[width=\textwidth]{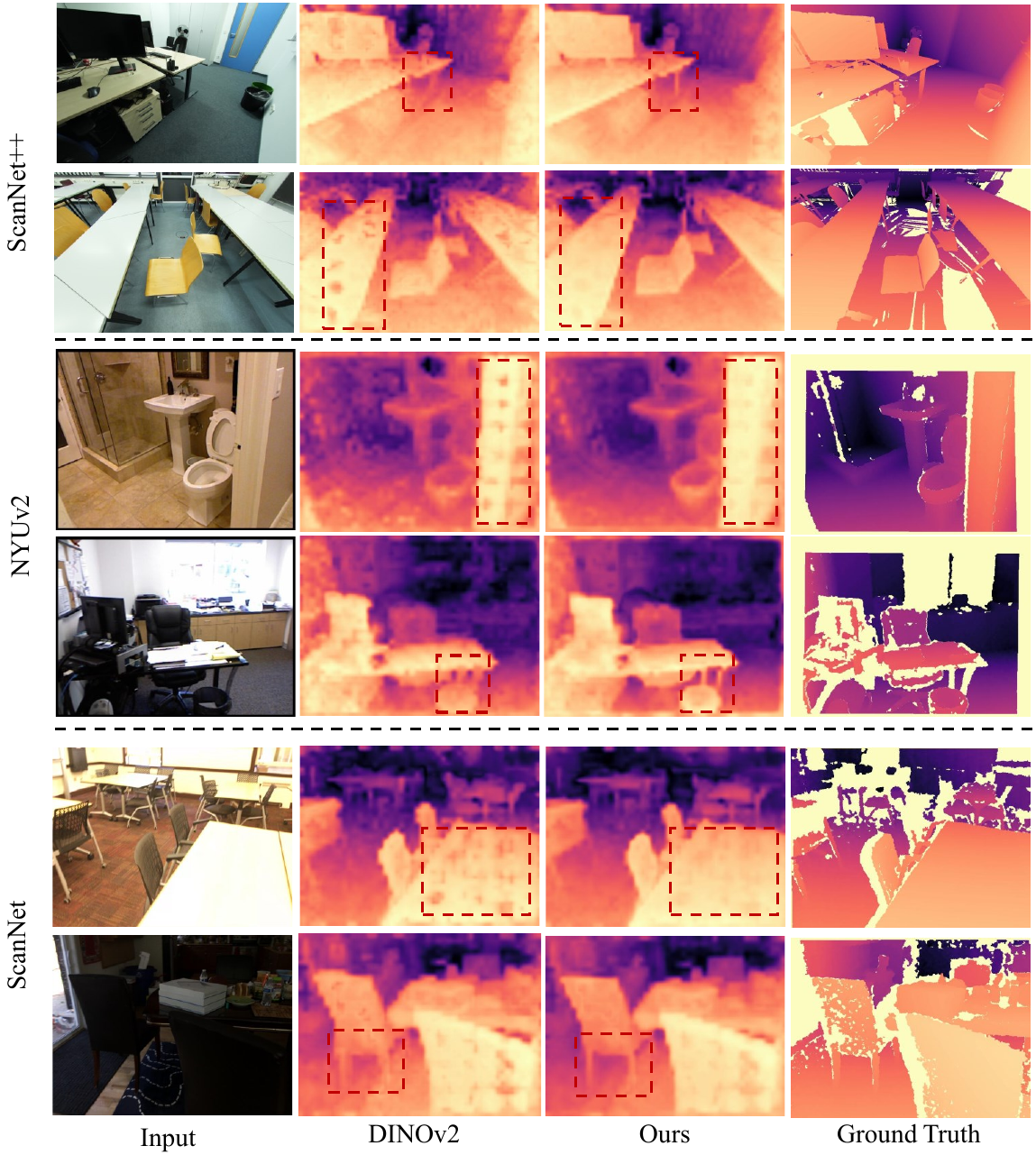}
    \caption{\small \textbf{Depth estimation on indoor datasets with linear probing}. Incorporating our 3D-aware fine-tuned features helps obtain cleaner depth in textureless regions and more detailed depth on fine-grained structures, \eg legs of tables or chairs.}
    \label{fig:indoor_depth}
\end{figure}

\vspace{0.1cm}
\noindent\textbf{Qualitative comparison.}
We qualitatively show the benefits of 3D-aware features in Fig.~\ref{fig:indoor_sem} and Fig.~\ref{fig:indoor_depth}. We observe the improvements are mainly reflected in two aspects: (1) cleaner segmentation/depth estimation in homogeneous or textureless regions, \eg on walls and boards, and (2) better prediction with fine-grained details, \eg on legs of chairs or tables. 
For (1), during the lifting of 2D features to 3D, features from multiple views are aggregated into a holistic representation, thus information from one view implicitly complements other views. 
We hypothesize that such multi-view awareness is transferred to DINOv2 through fine-tuning. 
By contrast, standard DINOv2 struggles to infer accurate segmentation or depth from a single image when with ambiguity, thus leading to noisy prediction. For (2), in our feature lifting process, we train the geometry properties (\eg position and opacity) of Gaussians with RGB color as supervision. The RGB guidance helps feature Gaussians learn detailed 3D structure and render high-resolution feature maps (\cf Fig.~\ref{fig:teaser} (c)). During the fine-tuning process, the model learns to estimate fine-grained features of objects (\cf Fig.~\ref{fig:teaser} (d) \vs (b)), which is helpful for capturing detailed structure in downstream tasks. 

\begin{table*}[ht]
\caption{
\textbf{Semantic segmentation scores on indoor datasets.} 3D-aware fine-tuning consistently leads to improved performance on semantic segmentation in comparison to standard DINOv2 across different indoor datasets.
}
\label{tab:indoor_sem}
\centering
\setlength{\tabcolsep}{5pt}
\resizebox{\textwidth}{!}{
\begin{tabular}{lccccccccc}
\toprule
\cmidrule{1-10} 
&
\multicolumn{3}{c}{ScanNet++~\cite{yeshwanth2023scannet++}} & \multicolumn{3}{c}{NYUv2~\cite{silberman2012indoor}} & \multicolumn{3}{c}{ScanNet~\cite{dai2017scannet}}\\
\cmidrule(r){2-4}	\cmidrule(r){5-7} \cmidrule(r){8-10}
Method  & mAcc ($\uparrow$) & mIoU ($\uparrow$) & aAcc ($\uparrow$) & mAcc ($\uparrow$) & mIoU ($\uparrow$) & aAcc ($\uparrow$)  & mAcc ($\uparrow$) & mIoU ($\uparrow$) & aAcc ($\uparrow$) \\
\midrule
DINOv2 \cite{oquab2023dinov2} & 40.84&30.19 & 80.25& 76.88& 65.55 & 82.43 &  55.86 & 43.6 & 73.54 \\
+ Ours & \textbf{43.4}& \textbf{32.76} & \textbf{83.54}  & \textbf{80.52} & \textbf{67.5} & \textbf{83.37} & \textbf{58.32}  & \textbf{44.84} & \textbf{74.37}\\ 
\bottomrule
\end{tabular}
}

\end{table*}

\vspace{-35px}
\begin{table*}[ht]
\caption{
\textbf{Depth estimation scores on indoor datasets.} 3D-aware fine-tuning consistently leads to improved performance on depth estimation in comparison to standard DINOv2 across different indoor datasets.
}
\label{tab:indoor_depth}
\centering
\setlength{\tabcolsep}{15pt}
\resizebox{\textwidth}{!}{
\begin{tabular}{lcccccc}
\toprule
\cmidrule{1-7} 
&
\multicolumn{2}{c}{ScanNet++~\cite{yeshwanth2023scannet++}} & \multicolumn{2}{c}{NYUv2~\cite{silberman2012indoor}} & \multicolumn{2}{c}{ScanNet~\cite{dai2017scannet}}\\
\cmidrule(r){2-3}	\cmidrule(r){4-5} \cmidrule(r){6-7}
Method  & RMSE ($\downarrow$) & Rel ($\downarrow$)&  RMSE ($\downarrow$)& Rel ($\downarrow$)&  RMSE ($\downarrow$)& Rel ($\downarrow$)\\
\midrule
DINOv2 \cite{oquab2023dinov2} & 0.3742&0.2836 & 0.4423&  0.1392 & 0.3089 & 0.1557 \\
+ Ours & \textbf{0.3361}& \textbf{0.2401} & \textbf{0.4198} & \textbf{0.1300} & \textbf{0.2921} & \textbf{0.1459} \\ 
\bottomrule
\end{tabular}
}

\end{table*}

\subsection{Out-of-domain Evaluation}
\label{sec:outdomain}
\begin{table*}[ht!]
\caption{
\textbf{Quantitative performance on out-of-domain datasets.}
3D-aware fine-tuning noticeably improves semantic segmentation on ADE20k and Pascal VOC and depth estimation on KITTI, demonstrating the transferability of the fine-tuned features, even under a significant domain gap.
}
\label{tab:out_of_domain}
\centering
\setlength{\tabcolsep}{5pt}
\resizebox{\textwidth}{!}{
\begin{tabular}{lcccccccc}
\toprule
\cmidrule{1-9} 
&
\multicolumn{3}{c}{ADE20k~\cite{zhou2017scene}} & \multicolumn{3}{c}{Pascal VOC~\cite{everingham2015pascal}} & \multicolumn{2}{c}{KITTI~\cite{geiger2013vision}}\\
\cmidrule(r){2-4}	\cmidrule(r){5-7} \cmidrule(r){8-9}
Method  & mAcc ($\uparrow$) & mIoU ($\uparrow$) & aAcc ($\uparrow$) & mAcc ($\uparrow$) & mIoU ($\uparrow$) & aAcc ($\uparrow$)  & RMSE ($\downarrow$)& Rel ($\downarrow$) \\
\midrule
DINOv2 \cite{oquab2023dinov2} & 56.74 & 44.28 & 79.73 &  90.61 & 81.14 & 95.72 & 3.03 & 0.10\\
+ Ours & \textbf{58.71} & \textbf{45.93} & \textbf{81.05}&  \textbf{91.04} &  \textbf{82.35} &  \textbf{96.14}  &  \textbf{2.91} & \textbf{0.09}\\ 
\bottomrule
\end{tabular}
}

\end{table*}

\begin{table}[ht]
\caption{
\textbf{Generalization on other 2D vision models.} Our 3D-aware fine-tuning applies to other 2D vision models and readily improves their performance.
}
\label{tab:other_models}
\setlength{\tabcolsep}{5pt}
\resizebox{\columnwidth}{!}{
\begin{tabular}{lcccccccc}
\toprule
\cmidrule{1-9} 
& \multicolumn{2}{c}{ \textbf{DINOv2-reg}}
 &
\multicolumn{2}{c}{ \textbf{CLIP}} & \multicolumn{2}{c}{ \textbf{MAE}} & \multicolumn{2}{c}{ \textbf{DeiT-III}} \\
\cmidrule(r){2-3}	\cmidrule(r){4-5} \cmidrule(r){6-7} \cmidrule(r){8-9}
  & mIoU ($\uparrow$)& RMSE ($\downarrow$)& mIoU ($\uparrow$)& RMSE ($\downarrow$)& mIoU ($\uparrow$)& RMSE ($\downarrow$) & mIoU ($\uparrow$)& RMSE ($\downarrow$)\\
\midrule
Original & 30.92 & 0.4190  & 25.61 & 0.4324 & 17.19 & 0.4855 & 18.62 & 0.4350 
\\
+ Ours & \textbf{33.39} & \textbf{0.3824} & \textbf{28.82} & \textbf{0.3960} &  \textbf{20.27} & \textbf{0.4795} & \textbf{22.98} & \textbf{0.3820} 
\\ 
\bottomrule
\end{tabular}
}
\end{table}

We train feature Gaussians and fine-tune DINOv2 on ScanNet++, a dataset that contains only indoor scenes with the usual content, \eg tables, chairs and other indoor furniture. We want to analyze how the gains obtained in this setting generalize to other domains, \eg outdoor scenes. For semantic segmentation, we conduct linear probing on ADE20k~\cite{zhou2017scene} and Pascal VOC~\cite{everingham2015pascal}. For depth estimation, we conduct linear probing on KITTI~\cite{geiger2013vision}.

\vspace{0.1cm}
\noindent\textbf{Quantitative comparison.} We observe that the improvement brought by 3D-aware features is generalizable to out-of-domain challenging datasets and also outdoor driving scenes, although to a smaller degree. As shown in Tab.~\ref{tab:out_of_domain}, for semantic segmentation task, incorporating our 3D-aware features brings a gain of 1.6\% mIoU on the   ADE20k~\cite{zhou2017scene} and a gain of 1.2\% mIoU on Pascal VOC~\cite{everingham2015pascal} over standard DINOv2 features. Furthermore, we also compare our performance on urban scene dataset KITTI~\cite{geiger2013vision} for depth estimation and observe our 3D-aware features help to reduce RMSE from 3.03 (DINOv2) to 2.91. %

\begin{figure}[ht]
\centering
\includegraphics[width=\textwidth]{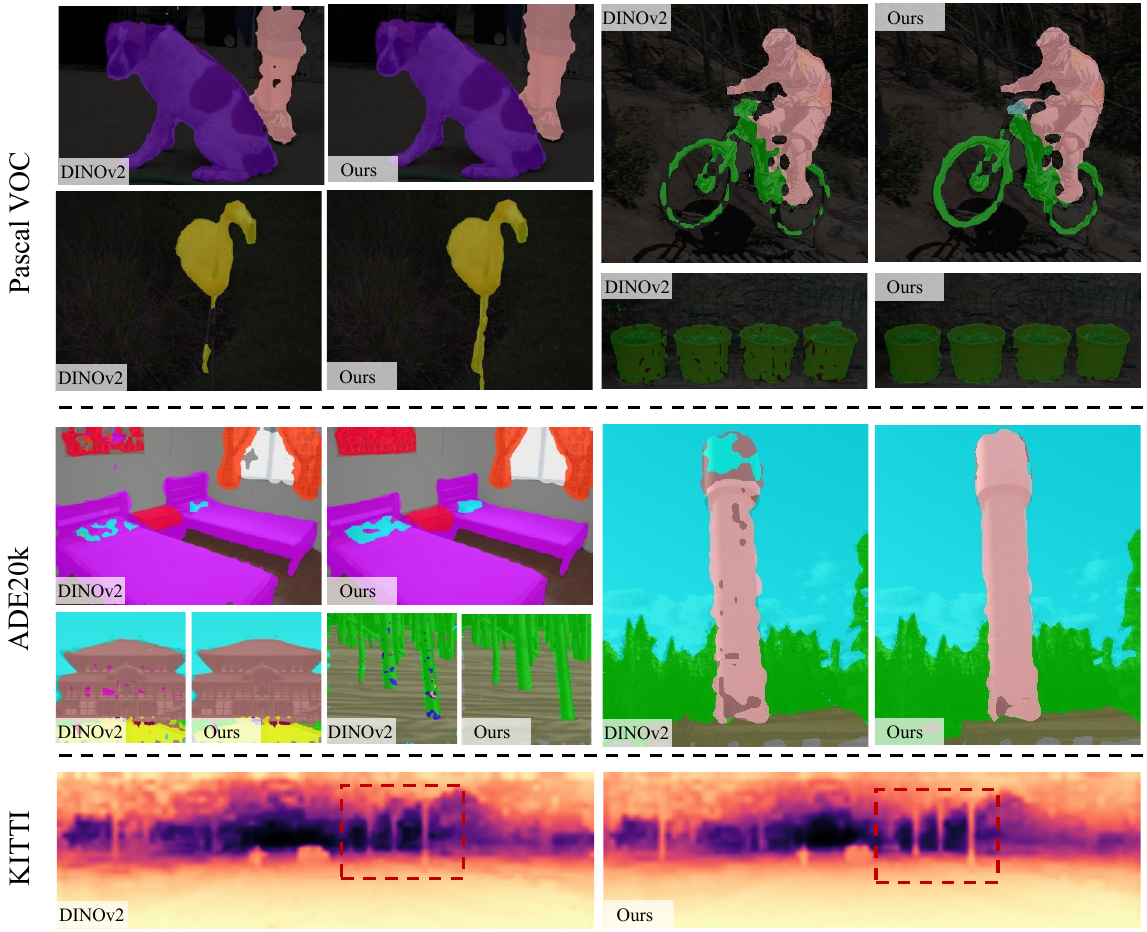}
    \caption{\small \textbf{Results on out-of-domain datasets with linear probing}. Our 3D-aware fine-tuned features help achieve better segmentation and capture detailed structure.}
    \label{fig:qualitative_out_of_domain}
\end{figure}

\vspace{0.1cm}
\noindent\textbf{Qualitative comparison.} We show qualitative comparison on out-of-domain datasets in Fig.~\ref{fig:qualitative_out_of_domain}. We observe similar improvements as in the within-domain datasets. Even though the 3D-aware fine-tuning is only conducted on indoor dataset ScanNet++, the fine-tuned features exhibit transferability to improve segmentation results for the detailed structures of common objects like bicycle and animal, and help achieve more compact segmentation of objects like tree, building and pillar. On depth estimation, the incorporated 3D-aware features are helpful in capturing the detailed structure of trees.

\subsection{Generalization to Other Vision Models}
\label{sec:multi_models}
We conduct experiments on more vision models (DINOv2-reg~\cite{darcet2023vision}, CLIP~\cite{radford2021learning}, MAE~\cite{he2022masked}, DeiT-III~\cite{touvron2022deit}) to prove the universality of our method. We show the linear probing results of semantic segmentation and depth estimation on ScanNet++ validation set in Tab.~\ref{tab:other_models}. Our method consistently improves all the models. We also visualize the features in Fig.~\ref{fig:addition_fig}. Note that there is little visual difference between the original MAE features and the fined-tuned features but our method still improves them.
\subsection{Ablation Studies and Analysis}
\label{subsec:ablation}
We conduct ablation studies on semantic segmentation on NYUv2 dataset with DINOv2.
\begin{table}[ht]
\begin{minipage}{.6\textwidth}
\caption{
\textbf{Ablation study on feature dimension of 3D Gaussian.} Increasing feature dimensions improves performance at the cost of larger memory consumption and longer training time.
}
\label{tab:ab_feature_dimension}
\centering
\setlength{\tabcolsep}{3pt}
\resizebox{\textwidth}{!}{
\begin{tabular}{cccccc}
\toprule
\cmidrule{1-6} 
Feature dimension & Average memory (MB) & Average Training time (h) & mAcc ($\uparrow$) & mIoU ($\uparrow$) & aAcc ($\uparrow$) \\
\midrule
32  & 370 & 1.3 & 78.77 & 67.15 & \textbf{83.44} \\
64 & 495 & 1.6 & \textbf{80.52}& \textbf{67.5} & 83.37  \\
128& 750 & 2.5 & - & -  & - \\
\bottomrule
\end{tabular}
}
\end{minipage}
\hspace{0.1cm}
\begin{minipage}{.38\textwidth}
\caption{
\textbf{Ablation study on fine-tuning \vs adapter.} An adapter is a tiny network plugged into the frozen DINOv2 features.
}
\label{tab:ab_adapter}
\centering
\setlength{\tabcolsep}{5pt}
\resizebox{\textwidth}{!}{
\begin{tabular}{cccc}
\toprule
\cmidrule{1-4}
Strategy &  mAcc ($\uparrow$) & mIoU ($\uparrow$) & aAcc ($\uparrow$) \\
\midrule
Fine-tuning  & \textbf{91.04} & \textbf{82.35} & \textbf{96.14}  \\
Adapter & 90.97 & 82.02 & 95.96  \\
\bottomrule
\end{tabular}
}
\end{minipage}
\end{table}

\begin{table}[ht]
\begin{minipage}{.45\textwidth}
\caption{
\textbf{Ablation study on feature assembly.} We study different strategies to assemble fine-tuned features with the original DINOv2.
}
\label{tab:ab_feature_assembly}
\centering
\setlength{\tabcolsep}{8pt}
\resizebox{\textwidth}{!}{
\begin{tabular}{cccc}
\toprule
\cmidrule{1-4} 
Strategy &  mAcc ($\uparrow$) & mIoU ($\uparrow$) & aAcc ($\uparrow$) \\
\midrule
Adding & 77.97 & 66.0 & 82.85  \\
Linear fusion & 78.22 & 66.39 & 82.89 \\
Concatenation   & \textbf{80.52}& \textbf{67.5} & \textbf{83.37} \\
\bottomrule
\end{tabular}
}
\end{minipage}
\hspace{0.1cm}
\begin{minipage}{.53\textwidth}
\caption{
\textbf{Ablation study on fine-tuning epochs.} We find fine-tuning with a single epoch with 8.5 hours is sufficient to achieve good performance.
}
\label{tab:ab_finetune_epochs}
\centering
\setlength{\tabcolsep}{5pt}
\resizebox{\textwidth}{!}{
\begin{tabular}{ccccc}
\toprule
\cmidrule{1-5}
Epochs &  Fine-tun. time (h) & mAcc ($\uparrow$) & mIoU ($\uparrow$) & aAcc ($\uparrow$) \\
\midrule
1   & 8.5 &  \textbf{80.52}& \textbf{67.5} & 83.37   \\
2 & 17 &  78.72 & 67.25 & \textbf{83.54}  \\
3 & 25.5 & 79.5 & 67.18 &  83.24 \\
\bottomrule
\end{tabular}
}
\end{minipage}
\end{table}

\vspace{0.1cm}
\noindent\textbf{Feature dimension of each Gaussian.} We attach a low-dimensional feature vector with each Gaussian and then up-project it to the same space with DINOv2. Tab.~\ref{tab:ab_feature_dimension} indicates that increasing the feature dimension from 32 to 64 will improve the performance of fine-tuned DINOv2 with an acceptable higher memory and longer training time. Increasing the feature dimension further to 128 is not feasible in our hardware due to the large memory consumption. We chose a feature dimension of 64 as a good compromise between model performance, memory consumption, as well
as training time.

\vspace{0.1cm}
\noindent\textbf{Feature assembly strategy.} We study different strategies to assemble the fine-tuned features with the original DINOv2 features in Tab.~\ref{tab:ab_feature_assembly}. We explore simple channel-wise adding and concatenation. Alternatively, we first concatenate the fine-tuned features with the original DINOv2 features then use a liner layer to fuse them to the same feature space of DINOv2. We observe simple concatenation works well in incorporating learned 3D-aware features while preserving original generalization ability.  

\vspace{0.1cm}
\noindent\textbf{Fine-tuning epochs.} We finetune DINOv2 using the features rendered by the pre-trained feature Gaussians. Tab.~\ref{tab:ab_finetune_epochs} suggests that a single epoch is sufficient to transfer the 3D awareness to DINOv2 and more epochs may harm the model's generalization ability.

\vspace{0.1cm}
\noindent\textbf{Fine-tuning \vs adapter.} Besides directly fine-tuning DINOv2, we explore an alternative strategy where we keep DINOv2 frozen and introduce an adapter on top of that. The adapter is a single Swin Transformer block~\cite{liu2021swin}. We observe the adapter can achieve comparable performance with fine-tuning (Tab.~\ref{tab:ab_adapter}), however, with longer training time. We stick with the fine-tuning strategy for simplicity without introducing any additional network component.

\vspace{0.1cm}
\noindent\textbf{Limitations and discussion.} 
Our work makes an initial step to transfer multi-view consistent and 3D-aware features encoded by a 3D Gaussian representation to 2D foundation model via fine-tuning. We demonstrate the 3D-aware features are helpful for downstream tasks. However, we still need the original features to keep the generalization ability. We attribute this to the limited diversity of our 3D training data (only a single indoor dataset) and hypothesize that this
can be remedied by fine-tuning on larger-scale data.

\section{Conclusion}
\label{sec:conclusion}

In this work, we present a method to inject 3D awareness into 2D foundation models. We first lift 2D foundation features into a 3D Gaussian representation and then use the rendered multi-view consistent and 3D-aware features to in turn fine-tune the 2D foundation model. Our experiments show that incorporating the fine-tuned features readily leads to improved performance on both semantic and geometric tasks through simple linear probing. Although we only conduct the 3D-aware fine-tuning on a single dataset ScanNet++, we demonstrate the learned 3D awareness is transferable across a variety of datasets in different domains. We hope our work inspires future research to consider equipping 2D foundation models with 3D-aware understanding.

\paragraph{Acknowledgements} Francis Engelmann is partially supported by an ETH AI Center postdoctoral research fellowship and an ETH Zurich Career Seed Award. This project was also partially supported by Saarbrücken Research Center for Visual Computing, Interaction and AI.

\bibliographystyle{splncs04}
\bibliography{egbib}

\begin{thebibliography}{10}
\providecommand{\url}[1]{\texttt{#1}}
\providecommand{\urlprefix}{URL }
\providecommand{\doi}[1]{https://doi.org/#1}

\bibitem{amir2021deep}
Amir, S., Gandelsman, Y., Bagon, S., Dekel, T.: {Deep ViT Features as Dense Visual Descriptors}. In: European Conference on Computer Vision (ECCV) Workshops (2022)

\bibitem{bachmann2022multimae}
Bachmann, R., Mizrahi, D., Atanov, A., Zamir, A.: {MultiMAE: Multi-modal Multi-task Masked Autoencoders}. In: European Conference on Computer Vision (ECCV) (2022)

\bibitem{bao2021beit}
Bao, H., Dong, L., Piao, S., Wei, F.: {BEiT: BERT Pre-training of Image Transformers}. In: International Conference on Learning Representations (ICLR) (2022)

\bibitem{bengio2013representation}
Bengio, Y., Courville, A., Vincent, P.: {Representation Learning: A Review and New Perspectives}. Transactions on Pattern Analysis and Machine Intelligence (PAMI)  (2013)

\bibitem{bhat2021adabins}
Bhat, S.F., Alhashim, I., Wonka, P.: {Adabins: Depth Estimation Using Adaptive Bins}. In: International Conference on Computer Vision and Pattern Recognition (CVPR) (2021)

\bibitem{caron2020unsupervised}
Caron, M., Misra, I., Mairal, J., Goyal, P., Bojanowski, P., Joulin, A.: {Unsupervised Learning of Visual Features by Contrasting Cluster Assignments}. In: International Conference on Neural Information Processing Systems (NeurIPS) (2020)

\bibitem{caron2021emerging}
Caron, M., Touvron, H., Misra, I., J{\'e}gou, H., Mairal, J., Bojanowski, P., Joulin, A.: {Emerging Properties in Self-Supervised Vision Transformers}. In: International Conference on Computer Vision (ICCV) (2021)

\bibitem{chen2020generative}
Chen, M., Radford, A., Child, R., Wu, J., Jun, H., Luan, D., Sutskever, I.: {Generative Pretraining From Pixels}. In: International Conference on Machine Learning (ICML) (2020)

\bibitem{dai2017scannet}
Dai, A., Chang, A.X., Savva, M., Halber, M., Funkhouser, T., Nie{\ss}ner, M.: {ScanNet: Richly-Annotated 3D Reconstructions of Indoor Scenes}. In: International Conference on Computer Vision and Pattern Recognition (CVPR) (2017)

\bibitem{darcet2023vision}
Darcet, T., Oquab, M., Mairal, J., Bojanowski, P.: {Vision Transformers Need Registers}. In: International Conference on Learning Representations (ICLR) (2024)

\bibitem{devlin2018bert}
Devlin, J., Chang, M.W., Lee, K., Toutanova, K.: {BERT: Pre-training of Deep Bidirectional Transformers for Language Understanding}. NAACL  (2018)

\bibitem{doersch2015unsupervised}
Doersch, C., Gupta, A., Efros, A.A.: {Unsupervised Visual Representation Learning by Context Prediction}. In: International Conference on Computer Vision (ICCV) (2015)

\bibitem{dosovitskiy2020image}
Dosovitskiy, A., Beyer, L., Kolesnikov, A., Weissenborn, D., Zhai, X., Unterthiner, T., Dehghani, M., Minderer, M., Heigold, G., Gelly, S., et~al.: {An Image Is Worth 16x16 Words: Transformers for Image Recognition at Scale}. In: International Conference on Learning Representations (ICLR) (2020)

\bibitem{dosovitskiy2014discriminative}
Dosovitskiy, A., Springenberg, J.T., Riedmiller, M., Brox, T.: {Discriminative Unsupervised Feature Learning With Convolutional Neural Networks}. In: International Conference on Neural Information Processing Systems (NeurIPS) (2014)

\bibitem{el2024probing}
El~Banani, M., Raj, A., Maninis, K.K., Kar, A., Li, Y., Rubinstein, M., Sun, D., Guibas, L., Johnson, J., Jampani, V.: {Probing the 3D Awareness of Visual Foundation Models}. In: International Conference on Computer Vision and Pattern Recognition (CVPR) (2024)

\bibitem{engelmann2024opennerf}
Engelmann, F., Manhardt, F., Niemeyer, M., Tateno, K., Tombari, F.: {OpenNeRF: Open Set 3D Neural Scene Segmentation with Pixel-Wise Features and Rendered Novel Views}. In: International Conference on Learning Representations (ICLR) (2024)

\bibitem{everingham2015pascal}
Everingham, M., Eslami, S.A., Van~Gool, L., Williams, C.K., Winn, J., Zisserman, A.: {The Pascal Visual Object Classes Challenge: A Retrospective}. International Journal of Computer Vision  (2015)

\bibitem{fu2024featup}
Fu, S., Hamilton, M., Brandt, L., Feldman, A., Zhang, Z., Freeman, W.T.: {FeatUp: A Model-Agnostic Framework for Features at Any Resolution}. In: International Conference on Learning Representations (ICLR) (2024)

\bibitem{geiger2013vision}
Geiger, A., Lenz, P., Stiller, C., Urtasun, R.: {Vision Meets Robotics: The Kitti Dataset}. The International Journal of Robotics Research  (2013)

\bibitem{ghiasi2022scaling}
Ghiasi, G., Gu, X., Cui, Y., Lin, T.Y.: {Scaling Open-Vocabulary Image Segmentation With Image-Level Labels}. In: European Conference on Computer Vision (ECCV) (2022)

\bibitem{gidaris2018unsupervised}
Gidaris, S., Singh, P., Komodakis, N.: {Unsupervised Representation Learning by Predicting Image Rotations}. In: International Conference on Learning Representations (ICLR) (2018)

\bibitem{goel2023interactive}
Goel, R., Sirikonda, D., Saini, S., Narayanan, P.: {Interactive Segmentation of Radiance Fields}. In: International Conference on Computer Vision and Pattern Recognition (CVPR) (2023)

\bibitem{grill2020bootstrap}
Grill, J.B., Strub, F., Altch{\'e}, F., Tallec, C., Richemond, P., Buchatskaya, E., Doersch, C., Avila~Pires, B., Guo, Z., Gheshlaghi~Azar, M., et~al.: {Bootstrap Your Own Latent-a New Approach to Self-Supervised Learning}. In: International Conference on Neural Information Processing Systems (NeurIPS) (2020)

\bibitem{ha2022semantic}
Ha, H., Song, S.: {Semantic Abstraction: Open-world 3D Scene Understanding From 2D Vision-Language Models}. In: Conference on Robot Learning (CoRL) (2022)

\bibitem{hadsell2006dimensionality}
Hadsell, R., Chopra, S., LeCun, Y.: {Dimensionality Reduction by Learning an Invariant Mapping}. In: International Conference on Computer Vision and Pattern Recognition (CVPR) (2006)

\bibitem{he2022masked}
He, K., Chen, X., Xie, S., Li, Y., Doll{\'a}r, P., Girshick, R.: {Masked Autoencoders Are Scalable Vision Learners}. In: International Conference on Computer Vision and Pattern Recognition (CVPR) (2022)

\bibitem{he2020momentum}
He, K., Fan, H., Wu, Y., Xie, S., Girshick, R.: {Momentum Contrast for Unsupervised Visual Representation Learning}. In: International Conference on Computer Vision and Pattern Recognition (CVPR) (2020)

\bibitem{hou2023mask3d}
Hou, J., Dai, X., He, Z., Dai, A., Nie{\ss}ner, M.: {Mask3D: Pre-training 2D Vision Transformers by Learning Masked 3D Priors}. In: International Conference on Computer Vision and Pattern Recognition (CVPR) (2023)

\bibitem{hou2021pri3d}
Hou, J., Xie, S., Graham, B., Dai, A., Nie{\ss}ner, M.: {Pri3D: Can 3D Priors Help 2D Representation Learning?} In: International Conference on Computer Vision (ICCV) (2021)

\bibitem{Huang2023Segment3D}
Huang, R., Peng, S., Takmaz, A., Tombari, F., Pollefeys, M., Song, S., Huang, G., Engelmann, F.: {Segment3D: Learning Fine-Grained Class-Agnostic 3D Segmentation without Manual Labels}. European Conference on Computer Vision (ECCV)  (2024)

\bibitem{jatavallabhula2023conceptfusion}
Jatavallabhula, K.M., Kuwajerwala, A., Gu, Q., Omama, M., Chen, T., Li, S., Iyer, G., Saryazdi, S., Keetha, N., Tewari, A., et~al.: {ConceptFusion: Open-set Multimodal 3D Mapping}. Robotics: Science and Systems (RSS)  (2023)

\bibitem{ke2023repurposing}
Ke, B., Obukhov, A., Huang, S., Metzger, N., Daudt, R.C., Schindler, K.: {Repurposing Diffusion-Based Image Generators for Monocular Depth Estimation}. In: International Conference on Computer Vision and Pattern Recognition (CVPR) (2024)

\bibitem{kerbl20233d}
Kerbl, B., Kopanas, G., Leimk{\"u}hler, T., Drettakis, G.: {3D Gaussian Splatting for Real-Time Radiance Field Rendering}. ACM Transactions on Graphics  (2023)

\bibitem{kerr2023lerf}
Kerr, J., Kim, C.M., Goldberg, K., Kanazawa, A., Tancik, M.: {LERF: Language Embedded Radiance Fields}. In: International Conference on Computer Vision (ICCV) (2023)

\bibitem{kirillov2023segment}
Kirillov, A., Mintun, E., Ravi, N., Mao, H., Rolland, C., Gustafson, L., Xiao, T., Whitehead, S., Berg, A.C., Lo, W.Y., et~al.: {Segment Anything}. In: International Conference on Computer Vision (ICCV) (2023)

\bibitem{kobayashi2022decomposing}
Kobayashi, S., Matsumoto, E., Sitzmann, V.: {Decomposing Nerf for Editing via Feature Field Distillation}. In: International Conference on Neural Information Processing Systems (NeurIPS) (2022)

\bibitem{li2022languagedriven}
Li, B., Weinberger, K.Q., Belongie, S., Koltun, V., Ranftl, R.: {Language-driven Semantic Segmentation}. In: International Conference on Learning Representations (ICLR) (2022)

\bibitem{li2023mask}
Li, F., Zhang, H., Xu, H., Liu, S., Zhang, L., Ni, L.M., Shum, H.Y.: {Mask DINO: Towards a Unified Transformer-Based Framework for Object Detection and Segmentation}. In: International Conference on Computer Vision and Pattern Recognition (CVPR) (2023)

\bibitem{liu2021swin}
Liu, Z., Lin, Y., Cao, Y., Hu, H., Wei, Y., Zhang, Z., Lin, S., Guo, B.: {Swin Transformer: Hierarchical Vision Transformer Using Shifted Windows}. In: International Conference on Computer Vision (ICCV) (2021)

\bibitem{loshchilov2017decoupled}
Loshchilov, I., Hutter, F.: {Decoupled Weight Decay Regularization}. In: International Conference on Learning Representations (ICLR) (2019)

\bibitem{mazur2022feature}
Mazur, K., Sucar, E., Davison, A.J.: {Feature-realistic Neural Fusion for Real-time, Open Set Scene Understanding}. In: International Conference on Robotics and Automation (ICRA) (2023)

\bibitem{mildenhall2021nerf}
Mildenhall, B., Srinivasan, P.P., Tancik, M., Barron, J.T., Ramamoorthi, R., Ng, R.: {NeRF: Representing Scenes as Neural Radiance Fields for View Synthesis}. In: European Conference on Computer Vision (ECCV) (2020)

\bibitem{noroozi2016unsupervised}
Noroozi, M., Favaro, P.: {Unsupervised Learning of Visual Representations by Solving Jigsaw Puzzles}. In: European Conference on Computer Vision (ECCV) (2016)

\bibitem{oquab2023dinov2}
Oquab, M., Darcet, T., Moutakanni, T., Vo, H., Szafraniec, M., Khalidov, V., Fernandez, P., Haziza, D., Massa, F., El-Nouby, A., et~al.: {DINOv2: Learning Robust Visual Features Without Supervision}. Transactions on Machine Learning Research  (2023)

\bibitem{pathak2017learning}
Pathak, D., Girshick, R., Doll{\'a}r, P., Darrell, T., Hariharan, B.: {Learning Features by Watching Objects Move}. In: International Conference on Computer Vision and Pattern Recognition (CVPR) (2017)

\bibitem{peng2022openscene}
Peng, S., Genova, K., Jiang, C., Tagliasacchi, A., Pollefeys, M., Funkhouser, T.: {OpenScene: 3D Scene Understanding with Open Vocabularies}. In: International Conference on Computer Vision and Pattern Recognition (CVPR) (2023)

\bibitem{Qin_2024_CVPR}
Qin, M., Li, W., Zhou, J., Wang, H., Pfister, H.: {LangSplat: 3D Language Gaussian Splatting}. In: International Conference on Computer Vision and Pattern Recognition (CVPR) (2024)

\bibitem{radford2021learning}
Radford, A., Kim, J.W., Hallacy, C., Ramesh, A., Goh, G., Agarwal, S., Sastry, G., Askell, A., Mishkin, P., Clark, J., et~al.: {Learning Transferable Visual Models From Natural Language Supervision}. In: International Conference on Machine Learning (ICML) (2021)

\bibitem{ranftl2021vision}
Ranftl, R., Bochkovskiy, A., Koltun, V.: {Vision Transformers for Dense Prediction}. In: International Conference on Computer Vision (ICCV) (2021)

\bibitem{rombach2022high}
Rombach, R., Blattmann, A., Lorenz, D., Esser, P., Ommer, B.: {High-Resolution Image Synthesis With Latent Diffusion Models}. In: International Conference on Computer Vision and Pattern Recognition (CVPR) (2022)

\bibitem{russakovsky2015imagenet}
Russakovsky, O., Deng, J., Su, H., Krause, J., Satheesh, S., Ma, S., Huang, Z., Karpathy, A., Khosla, A., Bernstein, M., et~al.: {Imagenet large scale visual recognition challenge}. International journal of computer vision  (2015)

\bibitem{saxena2024surprising}
Saxena, S., Herrmann, C., Hur, J., Kar, A., Norouzi, M., Sun, D., Fleet, D.J.: {The Surprising Effectiveness of Diffusion Models for Optical Flow and Monocular Depth Estimation}. In: International Conference on Neural Information Processing Systems (NeurIPS) (2024)

\bibitem{shen2023distilled}
Shen, W., Yang, G., Yu, A., Wong, J., Kaelbling, L.P., Isola, P.: {Distilled Feature Fields Enable Few-Shot Language-Guided Manipulation}. In: Conference on Robot Learning (CoRL) (2023)

\bibitem{shi2024language}
Shi, J.C., Wang, M., Duan, H.B., Guan, S.H.: {Language Embedded 3D Gaussians for Open-Vocabulary Scene Understanding}. In: International Conference on Computer Vision and Pattern Recognition (CVPR) (2024)

\bibitem{silberman2012indoor}
Silberman, N., Hoiem, D., Kohli, P., Fergus, R.: {Indoor Segmentation and Support Inference From RGBD Images}. In: European Conference on Computer Vision (ECCV) (2012)

\bibitem{Takmaz2023NIPS}
Takmaz, A., Fedele, E., Sumner, R.W., Pollefeys, M., Tombari, F., Engelmann, F.: {OpenMask3D: Open-Vocabulary 3D Instance Segmentation}. In: International Conference on Neural Information Processing Systems (NeurIPS) (2023)

\bibitem{tan2022semantic}
Tan, H., Wu, S., Pi, J.: {Semantic Diffusion Network for Semantic Segmentation}. In: International Conference on Neural Information Processing Systems (NeurIPS) (2022)

\bibitem{touvron2022deit}
Touvron, H., Cord, M., J{\'e}gou, H.: {Deit III: Revenge of the ViT}. In: European Conference on Computer Vision (ECCV) (2022)

\bibitem{tschernezki2022neural}
Tschernezki, V., Laina, I., Larlus, D., Vedaldi, A.: {Neural Feature Fusion Fields: 3D Distillation of Self-Supervised 2D Image Representations}. In: International Conference on 3D Vision (3DV) (2022)

\bibitem{wang2015unsupervised}
Wang, X., Gupta, A.: {Unsupervised Learning of Visual Representations Using Videos}. In: International Conference on Computer Vision (ICCV) (2015)

\bibitem{Weder2024labelmaker}
Weder, S., Blum, H., Engelmann, F., Pollefeys, M.: {LabelMaker: Automatic Semantic Label Generation from RGB-D Trajectories}. In: International Conference on 3D Vision (3DV) (2024)

\bibitem{weinzaepfel2022croco}
Weinzaepfel, P., Leroy, V., Lucas, T., Br{\'e}gier, R., Cabon, Y., Arora, V., Antsfeld, L., Chidlovskii, B., Csurka, G., Revaud, J.: {CroCo: Self-Supervised Pre-training for 3D Vision Tasks by Cross-View Completion}. In: International Conference on Neural Information Processing Systems (NeurIPS) (2022)

\bibitem{yang2024denoising}
Yang, J., Luo, K.Z., Li, J., Weinberger, K.Q., Tian, Y., Wang, Y.: {Denoising Vision Transformers}. In: European Conference on Computer Vision (ECCV) (2024)

\bibitem{yang2024depth}
Yang, L., Kang, B., Huang, Z., Xu, X., Feng, J., Zhao, H.: {Depth Anything: Unleashing the Power of Large-Scale Unlabeled Data}. In: International Conference on Computer Vision and Pattern Recognition (CVPR) (2024)

\bibitem{yeshwanth2023scannet++}
Yeshwanth, C., Liu, Y.C., Nie{\ss}ner, M., Dai, A.: {ScanNet++: A High-Fidelity Dataset of 3D Indoor Scenes}. In: International Conference on Computer Vision (ICCV) (2023)

\bibitem{zhang2024tale}
Zhang, J., Herrmann, C., Hur, J., Polania~Cabrera, L., Jampani, V., Sun, D., Yang, M.H.: {A Tale of Two Features: Stable Diffusion Complements DINO for Zero-Shot Semantic Correspondence}. In: International Conference on Neural Information Processing Systems (NeurIPS) (2023)

\bibitem{zhou2017scene}
Zhou, B., Zhao, H., Puig, X., Fidler, S., Barriuso, A., Torralba, A.: {Scene Parsing Through ade20K Dataset}. In: International Conference on Computer Vision and Pattern Recognition (CVPR) (2017)

\bibitem{zhou2024feature}
Zhou, S., Chang, H., Jiang, S., Fan, Z., Zhu, Z., Xu, D., Chari, P., You, S., Wang, Z., Kadambi, A.: {Feature 3DGS: Supercharging 3D Gaussian Splatting to Enable Distilled Feature Fields}. In: International Conference on Computer Vision and Pattern Recognition (CVPR) (2024)

\end{thebibliography}

\appendix
In the appendix, we provide (1) experiments with other DINOv2 ViT variants (Appendix~\ref{sec:dino_base}) (2) experiments on more tasks and heads (Appendix~\ref{sec:more_tasks_heads}) (3) experiments on impact of feature dimensions for linear probing (Appendix~\ref{sec:more_feat_dim}) (4) more visualization and K-Means clustering of features (Appendix~\ref{sec:features}).

\section{Experiments With More DINOv2 ViT Variants}
\label{sec:dino_base}

To demonstrate that the effectiveness of our 3D-aware fine-tuning is agnostic to DINOv2 architecture variants, we conduct additional experiments using the ViT-Base architecture with a feature dimension of 768. We show the results of semantic segmentation and depth estimation across multiple in-domain and out-of-domain datasets in Tab.~\ref{tab:supp_indoor_sem} and Tab.~\ref{tab:supp_depth}, respectively. We observe a similar trend of improvement with the ViT-B architecture. For example, on the fine-tuning dataset ScanNet++, incorporating our fine-tuned features brings an improvement of 3.47\% mIoU on semantic segmentation and reduces 0.03 RMSE on depth estimation. On other indoor datasets NYUv2 and out-of-domain dataset ADE20k, our 3D-aware fine-tuning consistently improves the original DINOv2. This experiment indicates that our 3D-aware fine-tuning is applicable to different ViT architectures and readily benefits downstream tasks.

\begin{table*}[ht]
\caption{
\textbf{Results of ViT variants on semantic segmentation.} Our 3D-aware fine-tuning yields consistent improvements on semantic segmentation for both ViT-S and ViT-B architectures.
}
\label{tab:supp_indoor_sem}
\centering
\setlength{\tabcolsep}{5pt}
\resizebox{\textwidth}{!}{
\begin{tabular}{lcccccccccc}
\toprule
\cmidrule{1-11} 
&
&
\multicolumn{3}{c}{ScanNet++~\cite{yeshwanth2023scannet++}} & \multicolumn{3}{c}{NYUv2~\cite{silberman2012indoor}} & \multicolumn{3}{c}{ADE20k~\cite{zhou2017scene}}\\
\cmidrule(r){3-5}	\cmidrule(r){6-8} \cmidrule(r){9-11}
Method &  Arch. & mAcc ($\uparrow$) & mIoU ($\uparrow$) & aAcc ($\uparrow$) & mAcc ($\uparrow$) & mIoU ($\uparrow$) & aAcc ($\uparrow$)  & mAcc ($\uparrow$) & mIoU ($\uparrow$) & aAcc ($\uparrow$) \\
\midrule
DINOv2 \cite{oquab2023dinov2} & ViT-S & 40.84&30.19 & 80.25& 76.88& 65.55 & 82.43 &  56.74 & 44.28 &79.73 \\
+ Ours & ViT-S & \textbf{43.4}& \textbf{32.76} & \textbf{83.54}  & \textbf{80.52} & \textbf{67.5} & \textbf{83.37} & \textbf{58.71}  & \textbf{45.93} & \textbf{81.05}\\ 

\midrule

DINOv2 \cite{oquab2023dinov2} & ViT-B & 42.99 & 32.72 & 82.05  & 80.56 & 68.45 & 84.03 & 59.11& 47.16& 80.79\\
+ Ours & ViT-B &  \textbf{46.35} &  \textbf{36.19} &  \textbf{85.5} & \textbf{80.58} & \textbf{70.56}& \textbf{85.72}& \textbf{62.18} & \textbf{49.5}& \textbf{82.52}\\ 
\bottomrule
\end{tabular}
}

\end{table*}

\begin{table*}[ht]
\caption{
\textbf{Results of ViT variants on depth estimation.} Our 3D-aware fine-tuning yields consistent improvements on depth segmentation for both ViT-S and ViT-B architectures.
}
\label{tab:supp_depth}
\centering
\setlength{\tabcolsep}{5pt}
\resizebox{\textwidth}{!}{
\begin{tabular}{lccccccc}
\toprule
\cmidrule{1-8} 
& &
\multicolumn{2}{c}{ScanNet++~\cite{yeshwanth2023scannet++}} & \multicolumn{2}{c}{NYUv2~\cite{silberman2012indoor}} & \multicolumn{2}{c}{KITTI~\cite{geiger2013vision}}\\
\cmidrule(r){3-4}	\cmidrule(r){5-6} \cmidrule(r){7-8}
Method  & Arch. & RMSE ($\downarrow$) & Rel ($\downarrow$)&  RMSE ($\downarrow$)& Rel ($\downarrow$)&  RMSE ($\downarrow$)& Rel ($\downarrow$)\\
\midrule
DINOv2 \cite{oquab2023dinov2} & ViT-S& 0.3742&0.2836 & 0.4423&  0.1392 & 3.0322 & 0.0965 \\
+ Ours & ViT-S & \textbf{0.3361}& \textbf{0.2401} & \textbf{0.4198} & \textbf{0.1300} & \textbf{2.9125} & \textbf{0.0891} \\ 
\midrule
DINOv2 \cite{oquab2023dinov2} & ViT-B& 0.3439 & 0.2576 & 0.3986 & 0.1218 & 2.9071 & 0.095 \\
+ Ours & ViT-B & \textbf{0.3174} & \textbf{0.2324} & \textbf{0.3802} & \textbf{0.1171} & \textbf{2.7923} & \textbf{0.0897}\\ 

\bottomrule
\end{tabular}
}

\end{table*}

\section{Experiments on More Tasks and Heads}
\label{sec:more_tasks_heads}
\begin{table}[h]
\begin{minipage}{.4\textwidth}
\caption{
\textbf{Results on image classification.} Our features do not improve image classification results.
}
\label{tab:img_cls}
\centering
\setlength{\tabcolsep}{15pt}
\resizebox{\textwidth}{!}{
\begin{tabular}{cc}
\toprule
\cmidrule{1-2} 
Method & Acc. ($\uparrow$) \\
\midrule
DINOv2~\cite{oquab2023dinov2}  &  \textbf{80.02} \\
+ Ours & 80.00  \\
\bottomrule
\end{tabular}
}
\end{minipage}
\hspace{0.1cm}
\begin{minipage}{.58\textwidth}
\caption{
\textbf{Results with DPT head on depth estimation.} Beyond linear probing, we evaluate with DPT head for depth estimation and observe consistent improvement.
}
\label{tab:dpt_head}
\centering
\setlength{\tabcolsep}{15pt}
\resizebox{\textwidth}{!}{
\begin{tabular}{ccc}
\toprule
\cmidrule{1-3}
Method &  RMSE ($\downarrow$) & Rel ($\downarrow$)  \\
\midrule
DINOv2~\cite{oquab2023dinov2}  & 0.3027 & 0.2149  \\
+ Ours & \textbf{0.2830} & \textbf{0.1936} \\
\bottomrule
\end{tabular}
}
\end{minipage}
\end{table}

\vspace{0.2 cm}
\noindent\textbf{Image classification.} We additionally evaluate our approach with DINOv2 small on image classification. We train a linear probing on ImageNet-1K~\cite{russakovsky2015imagenet} for 12500 iterations on a single GPU. As shown in tab.~\ref{tab:img_cls}, our features do not improve image classification results. This is expected as classification mainly relies on \texttt{CLS} token of ViT while our method aims to improve image patch features. 

\vspace{0.2 cm}
\noindent\textbf{DPT head.} Beyond linear probing, we evaluate DINOv2 small with the DPT head~\cite{ranftl2021vision} for depth estimation on ScanNet++. In comparison with the linear probing results (Tab.~2 in the main paper), the DPT head improves both results and our features are still helpful in this setup (see Tab.~\ref{tab:dpt_head}). This demonstrates that improvement brought the 3D-aware features is not limited to linear probing but also applicable to more complex heads.

\section{Experiments on Feature Dimensions}
\label{sec:more_feat_dim}
\begin{table*}[ht]
\caption{
\textbf{Results of duplicating DINOv2 features for linear probing.} We verify that it is not the number of feature dimensions that leads to improvement by showing that simple duplication of original features does not help.
}
\label{tab:feat_dim}
\centering
\setlength{\tabcolsep}{20pt}
\resizebox{\textwidth}{!}{
\begin{tabular}{llll}
\toprule
\cmidrule{1-4} 
 & $F_{dim}$ & mIoU ($\uparrow$)& RMSE ($\downarrow$)\\
\midrule
\circlenum{1} DINOv2 & 384 & 30.19 & 0.3742
\\
\circlenum{2} DINOv2 × 2  & 768 & 30.31 & 0.3676
\\
\circlenum{3} DINOv2 + Ours  & 768 & \textbf{32.76} & \textbf{0.3361}
\\ 
\bottomrule
\end{tabular}
}

\end{table*}

We concatenate original 2D features with our fine-tuned features, which will introduce increased feature dimension. In this experiment, we compare with DINOv2 small with duplicate features for linear probing of semantic segmentation and depth estimation on ScanNet++. As shown in Tab.~\ref{tab:feat_dim}, simply duplication \circlenum{2} only leads to little improvement compared with incorporating our fine-tuned features \circlenum{3}. This verifies that it is not the number of feature dimensions that leads to improvement. 

\section{Visual Analysis of Features}
\label{sec:features}

\begin{figure}
\centering
\includegraphics[width=\textwidth]{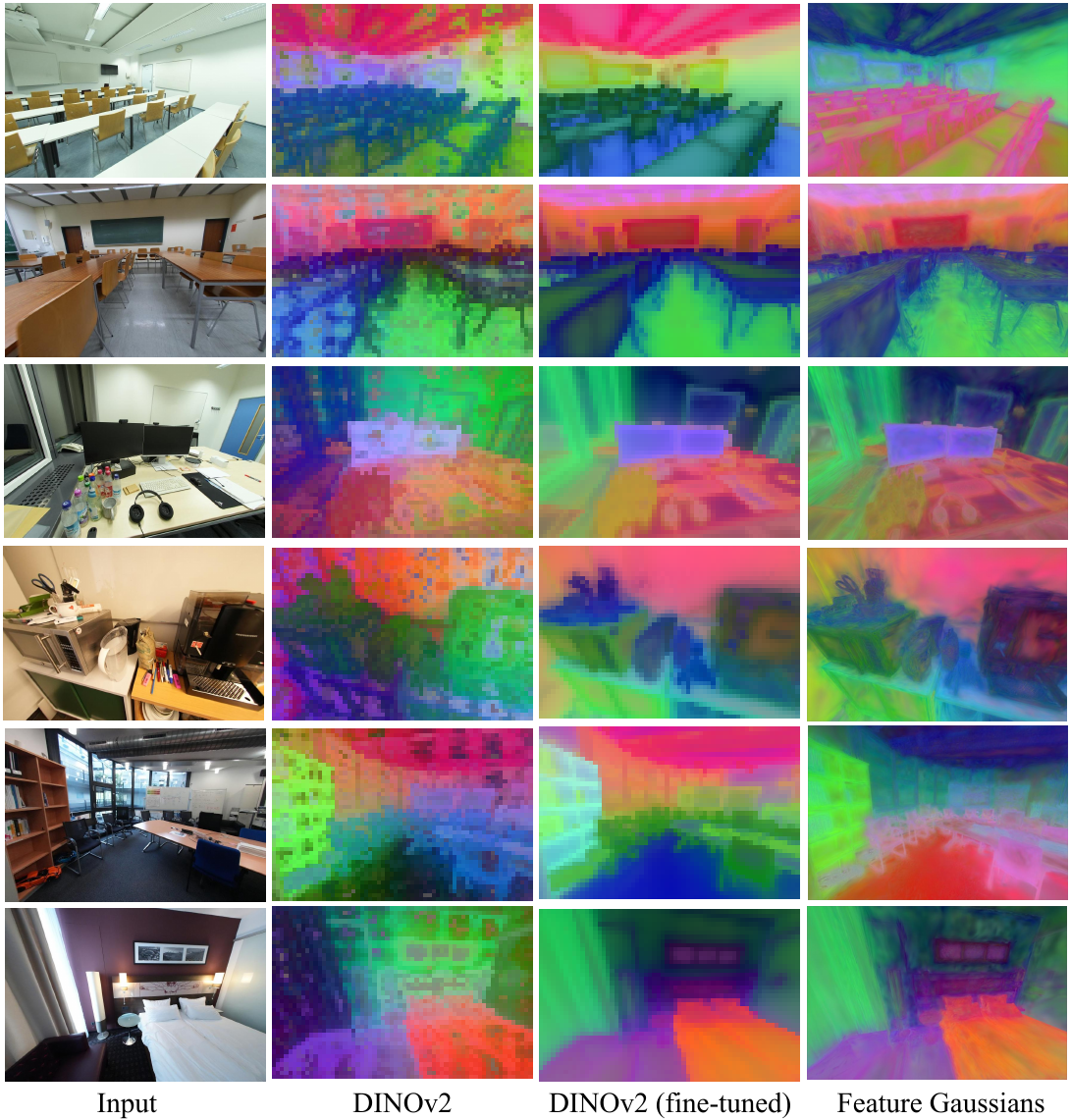}
    \caption{\small \textbf{Feature visualization  on ScanNet++~\cite{yeshwanth2023scannet++}}. We visualize the features rendered by pre-trained feature Gaussians (4\textsuperscript{th} column), features of DINOv2 (2\textsuperscript{nd} column) and our fine-tuned features (3\textsuperscript{rd} column). 
    }
    \label{fig:features_scannetpp}
\end{figure}

We train feature Gaussians and conduct 3D-aware fine-tuning on ScanNet++. In Fig~\ref{fig:features_scannetpp}, we visualize the features rendered by pre-trained feature Gaussians (4\textsuperscript{th} column), features of DINOv2 (2\textsuperscript{nd} column) and our fine-tuned features (3\textsuperscript{rd} column). 
The colors of features in all visualizations are produced using principle component analysis (PCA).
The standard DINOv2 features suffer from noise and rough object boundaries. After lifting those features to 3D by training feature Gaussians, we observe the rendered features enjoy cleaner and sharper object boundaries. We then fine-tune DINOv2 using those rendered features, which results in compact and clean feature representations.

\begin{figure}
\centering
\includegraphics[width=1\textwidth]{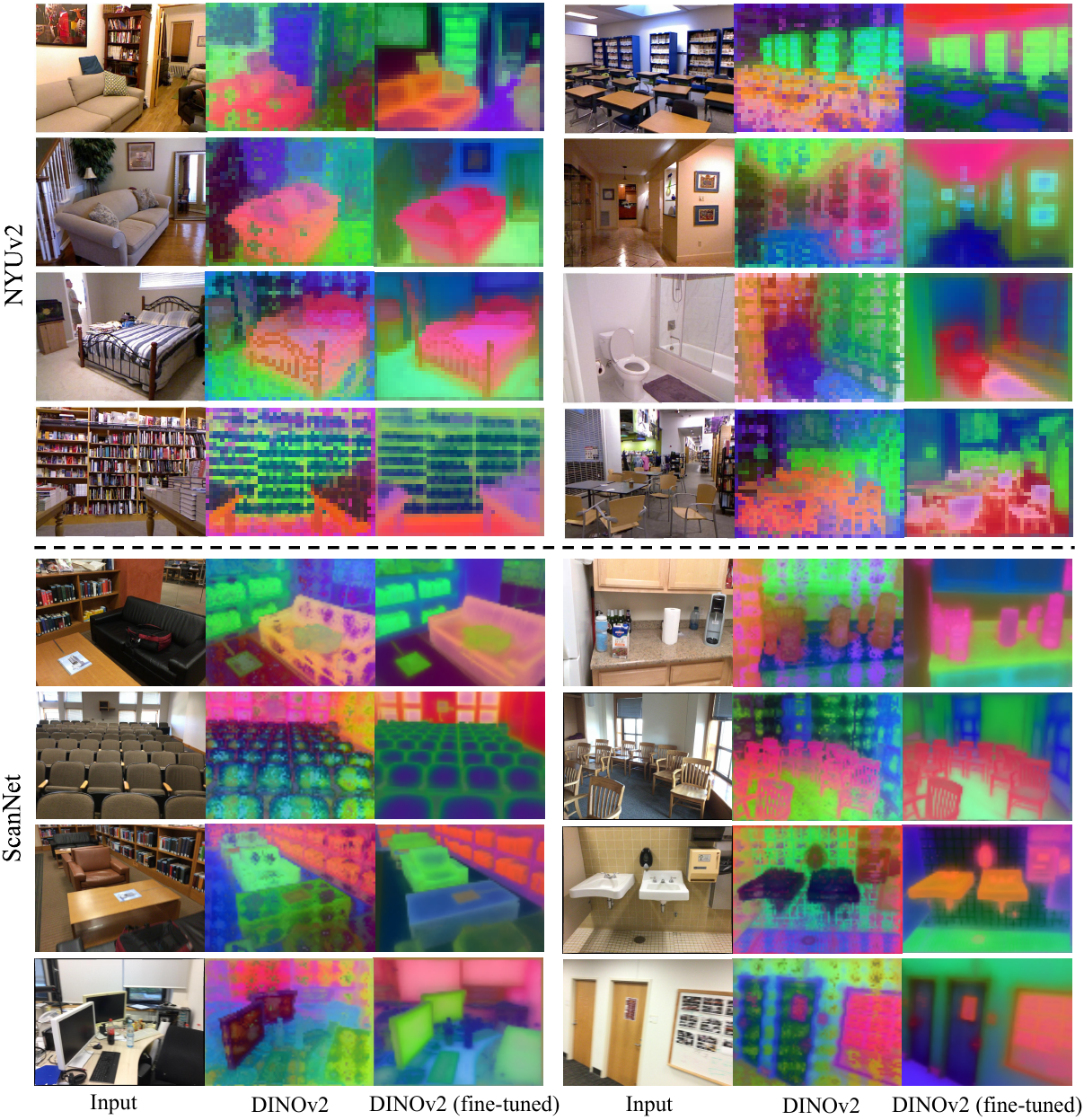}
    \caption{\small \textbf{Feature visualization on indoor datasets NYUv2~\cite{silberman2012indoor} and ScanNet~\cite{dai2017scannet}}. Our 3D-aware fine-tuning helps obtain more compact features and capture detailed structures.}
    \label{fig:features_indoor}
\end{figure}

\begin{figure}
\centering
\includegraphics[width=1\textwidth]{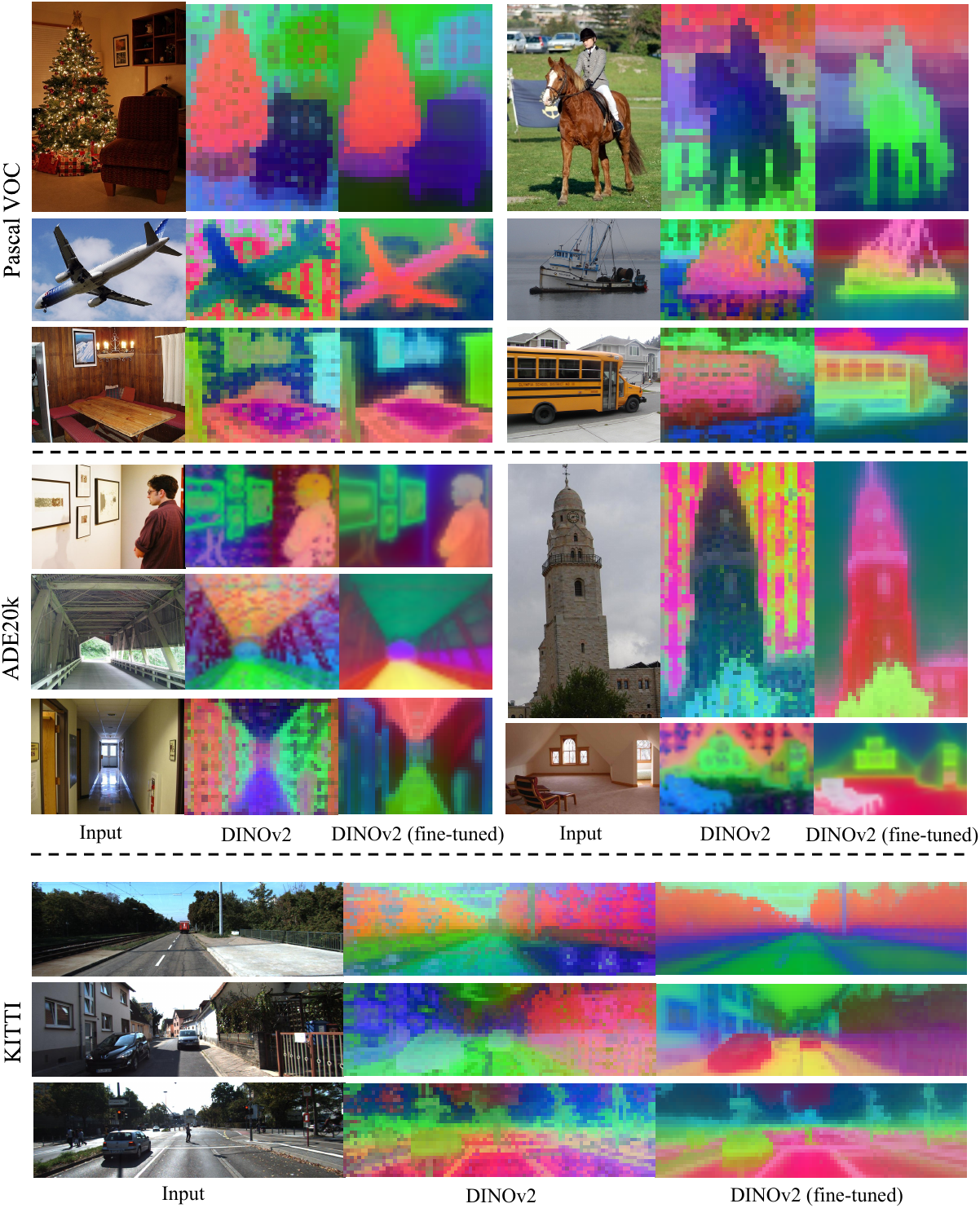}
    \caption{\small \textbf{Feature visualization  on out-of-domain datasets}. Our 3D-aware fine-tuning is generalizable to out-of-domain datasets and helps obtain more compact features and capture detailed structures.}
    \label{fig:features_out_domain}
\end{figure}

Although the fine-tuning is only conducted
on ScanNet++, we observe the resulting fine-tuned DINOv2 can generalize to other indoor datasets (\eg NYUv2 and ScanNet) and produces cleaner feature maps and more pronounced structure details (Fig.~\ref{fig:features_indoor}).  
Similar patterns can also be found in out-of-domain datasets (\eg Pascal VOC, ADE20k and KITTI), as shown in Fig.~\ref{fig:features_out_domain}. Visualizations of these feature representations indicate that 3D-aware fine-tuning is helpful and transferable. We observe the improvements are mainly reflected in two aspects: (1) cleaner and more compact feature maps.  (2) clearer object boundaries and structured details emerge.

\begin{figure}
\centering
\includegraphics[width=1\textwidth]{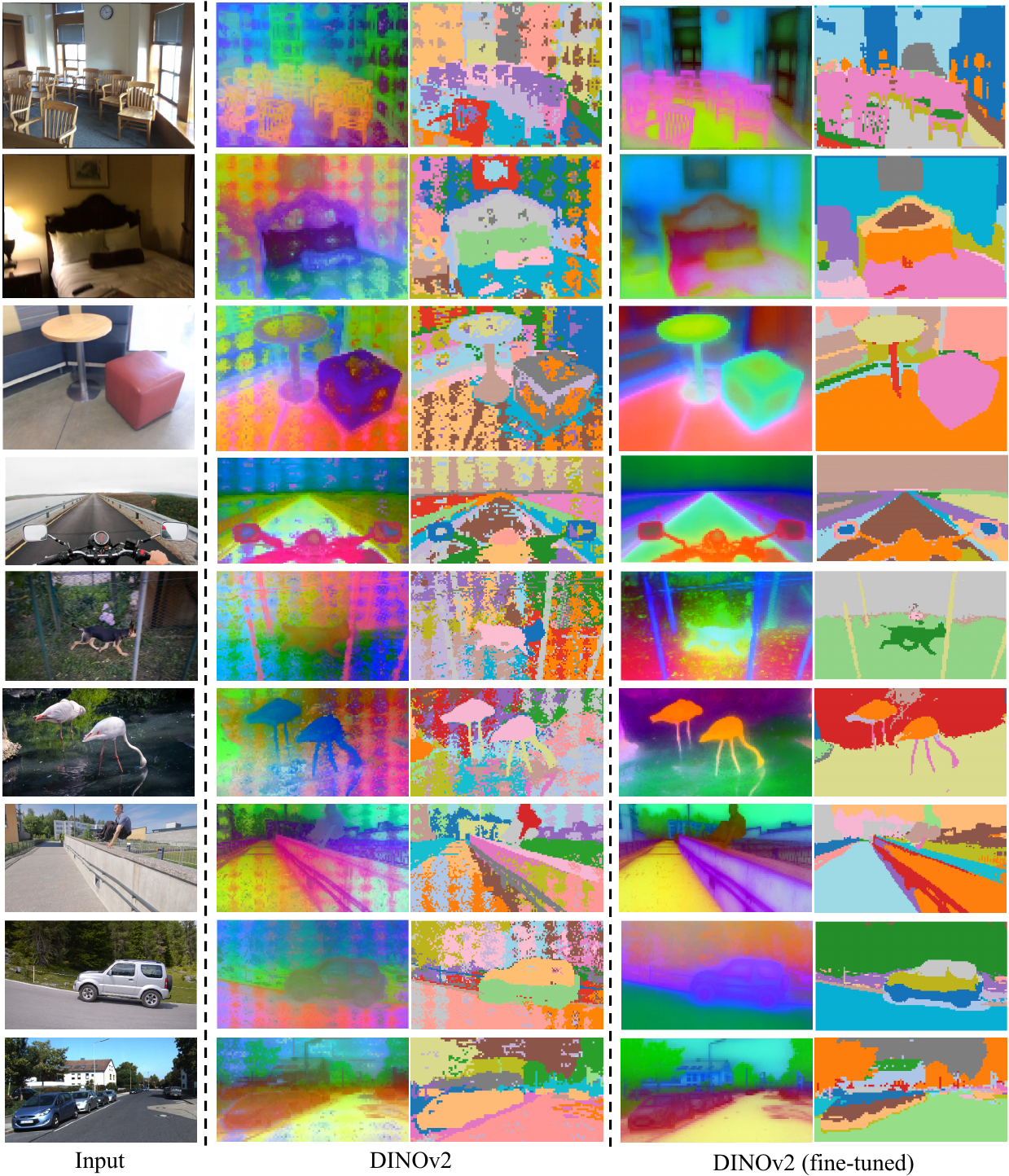}
    \caption{\small \textbf{K-Means clustering of features}. We show the PCA features and K-Means clustering results of DINOv2 (2, 3\textsuperscript{th} columns) and our 3D-aware fine-tuning features (4, 5\textsuperscript{th} columns). Our K-Means clustering results are more compact and detailed than DINOv2.}
    \label{fig:viz_kmeans}
\end{figure}

\vspace{0.2 cm}
\noindent\textbf{Feature clustering.}
We also use a simple K-Means clustering to directly examine the semantic concepts encoded in the feature representations. We show the K-means clustering results in Fig.~\ref{fig:viz_kmeans}. The improvements in our features are directly reflected in those simple clustering results. As shown in Fig.~\ref{fig:viz_kmeans}, the K-Means results of DINOv2 (3\textsuperscript{rd} column) are strongly affected by artifacts and noise. By contrast, our clustering results (5\textsuperscript{th} column) are much cleaner and more compact. In addition, we observe the PCA features and K-Means clustering of our 3D-aware fine-tuned features exhibit higher temporal consistency than the standard DINOv2 features. 
Please check our demos on our project page to see the full visualizations of video sequences.

\end{document}